\documentclass[runningheads]{llncs}
\usepackage{graphicx}
\usepackage{amsmath,amssymb} 
\usepackage{color}
\usepackage{booktabs}

\begin{document}
\title{Iterative Crowd Counting} 

\titlerunning{ic-CNN for Iterative Crowd Counting}

\authorrunning{Ranjan, Le, Hoai}

\authorrunning{Ranjan, Le, Hoai}
\author{Viresh Ranjan \and Hieu Le \and Minh Hoai}


\institute{Department of Computer Science,\\
	Stony Brook University\\
	\email{ \{vranjan,hle,minhhoai\}@cs.stonybrook.edu}
}

\def\mA{\mathcal{A}}
\def\mB{\mathcal{B}}
\def\mC{\mathcal{C}}
\def\mD{\mathcal{D}}
\def\mE{\mathcal{E}}
\def\mF{\mathcal{F}}
\def\mG{\mathcal{G}}
\def\mH{\mathcal{H}}
\def\mI{\mathcal{I}}
\def\mJ{\mathcal{J}}
\def\mK{\mathcal{K}}
\def\mL{\mathcal{L}}
\def\mM{\mathcal{M}}
\def\mN{\mathcal{N}}
\def\mO{\mathcal{O}}
\def\mP{\mathcal{P}}
\def\mQ{\mathcal{Q}}
\def\mR{\mathcal{R}}
\def\mS{\mathcal{S}}
\def\mT{\mathcal{T}}
\def\mU{\mathcal{U}}
\def\mV{\mathcal{V}}
\def\mW{\mathcal{W}}
\def\mX{\mathcal{X}}
\def\mY{\mathcal{Y}}
\def\mZ{\mathcal{Z}}

\def\1n{\mathbf{1}_n}
\def\0{\mathbf{0}}
\def\1{\mathbf{1}}

\def\A{{\bf A}}
\def\B{{\bf B}}
\def\C{{\bf C}}
\def\D{{\bf D}}
\def\E{{\bf E}}
\def\F{{\bf F}}
\def\G{{\bf G}}
\def\H{{\bf H}}
\def\I{{\bf I}}
\def\J{{\bf J}}
\def\K{{\bf K}}
\def\L{{\bf L}}
\def\M{{\bf M}}
\def\N{{\bf N}}
\def\O{{\bf O}}
\def\P{{\bf P}}
\def\Q{{\bf Q}}
\def\R{{\bf R}}
\def\S{{\bf S}}
\def\T{{\bf T}}
\def\U{{\bf U}}
\def\V{{\bf V}}
\def\W{{\bf W}}
\def\X{{\bf X}}
\def\Y{{\bf Y}}
\def\Z{{\bf Z}}

\def\a{{\bf a}}
\def\b{{\bf b}}
\def\c{{\bf c}}
\def\d{{\bf d}}
\def\e{{\bf e}}
\def\f{{\bf f}}
\def\g{{\bf g}}
\def\h{{\bf h}}
\def\i{{\bf i}}
\def\j{{\bf j}}
\def\k{{\bf k}}
\def\l{{\bf l}}
\def\m{{\bf m}}
\def\n{{\bf n}}
\def\o{{\bf o}}
\def\p{{\bf p}}
\def\q{{\bf q}}
\def\r{{\bf r}}
\def\s{{\bf s}}
\def\t{{\bf t}}
\def\u{{\bf u}}
\def\v{{\bf v}}
\def\w{{\bf w}}
\def\x{{\bf x}}
\def\y{{\bf y}}
\def\z{{\bf z}}

\def\balpha{\mbox{\boldmath{$\alpha$}}}
\def\bbeta{\mbox{\boldmath{$\beta$}}}
\def\bdelta{\mbox{\boldmath{$\delta$}}}
\def\bgamma{\mbox{\boldmath{$\gamma$}}}
\def\blambda{\mbox{\boldmath{$\lambda$}}}
\def\bsigma{\mbox{\boldmath{$\sigma$}}}
\def\btheta{\mbox{\boldmath{$\theta$}}}
\def\bomega{\mbox{\boldmath{$\omega$}}}
\def\bxi{\mbox{\boldmath{$\xi$}}}
\def\bnu{\mbox{\boldmath{$\nu$}}}                                  
\def\bphi{\mbox{\boldmath{$\phi$}}}
\def\bmu{\mbox{\boldmath{$\mu$}}}

\def\bDelta{\mbox{\boldmath{$\Delta$}}}
\def\bOmega{\mbox{\boldmath{$\Omega$}}}
\def\bPhi{\mbox{\boldmath{$\Phi$}}}
\def\bLambda{\mbox{\boldmath{$\Lambda$}}}
\def\bSigma{\mbox{\boldmath{$\Sigma$}}}
\def\bGamma{\mbox{\boldmath{$\Gamma$}}}

\newcommand{\myminimum}[1]{\mathop{\textrm{minimum}}_{#1}}
\newcommand{\mymaximum}[1]{\mathop{\textrm{maximum}}_{#1}}    
\newcommand{\mymin}[1]{\mathop{\textrm{minimize}}_{#1}}
\newcommand{\mymax}[1]{\mathop{\textrm{maximize}}_{#1}}
\newcommand{\mymins}[1]{\mathop{\textrm{min.}}_{#1}}
\newcommand{\mymaxs}[1]{\mathop{\textrm{max.}}_{#1}}  
\newcommand{\myargmin}[1]{\mathop{\textrm{argmin}}_{#1}} 
\newcommand{\myargmax}[1]{\mathop{\textrm{argmax}}_{#1}} 
\newcommand{\myst}{\textrm{s.t. }}

\newcommand{\denselist}{\itemsep -1pt}
\newcommand{\sparselist}{\itemsep 1pt}

\definecolor{pink}{rgb}{0.9,0.5,0.5}
\definecolor{purple}{rgb}{0.5, 0.4, 0.8}   
\definecolor{gray}{rgb}{0.3, 0.3, 0.3}
\definecolor{mygreen}{rgb}{0.2, 0.6, 0.2}

\newcommand{\cyan}[1]{\textcolor{cyan}{#1}}
\newcommand{\red}[1]{\textcolor{red}{#1}}  
\newcommand{\blue}[1]{\textcolor{blue}{#1}}
\newcommand{\magenta}[1]{\textcolor{magenta}{#1}}
\newcommand{\pink}[1]{\textcolor{pink}{#1}}
\newcommand{\green}[1]{\textcolor{green}{#1}} 
\newcommand{\gray}[1]{\textcolor{gray}{#1}}    
\newcommand{\mygreen}[1]{\textcolor{mygreen}{#1}}    
\newcommand{\purple}[1]{\textcolor{purple}{#1}}       

\definecolor{greena}{rgb}{0.4, 0.5, 0.1}
\newcommand{\greena}[1]{\textcolor{greena}{#1}}

\definecolor{bluea}{rgb}{0, 0.4, 0.6}
\newcommand{\bluea}[1]{\textcolor{bluea}{#1}}
\definecolor{reda}{rgb}{0.6, 0.2, 0.1}
\newcommand{\reda}[1]{\textcolor{reda}{#1}}

\def\changemargin#1#2{\list{}{\rightmargin#2\leftmargin#1}\item[]}
\let\endchangemargin=\endlist
                                               
\newcommand{\cm}[1]{}

\newcommand{\mtodo}[1]{{\color{red}$\blacksquare$\textbf{[TODO: #1]}}}
\newcommand{\myheading}[1]{\vspace{1ex}\noindent \textbf{#1}}
\newcommand{\htimesw}[2]{\mbox{$#1$$\times$$#2$}}


\newif\ifshowsolution
\showsolutiontrue

\ifshowsolution  
\newcommand{\Comment}[1]{\paragraph{\bf $\bigstar $ COMMENT:} {\sf #1} \bigskip}
\newcommand{\Solution}[2]{\paragraph{\bf $\bigstar $ SOLUTION:} {\sf #2} }
\newcommand{\Mistake}[2]{\paragraph{\bf $\blacksquare$ COMMON MISTAKE #1:} {\sf #2} \bigskip}
\else
\newcommand{\Solution}[2]{\vspace{#1}}
\fi

\newcommand{\truefalse}{
\begin{enumerate}
	\item True
	\item False
\end{enumerate}
}

\newcommand{\yesno}{
\begin{enumerate}
	\item Yes
	\item No
\end{enumerate}
}

\maketitle              
\begin{abstract}
   In this work, we tackle the problem of crowd counting in images. We present a Convolutional Neural Network (CNN) based density estimation approach to solve this problem. Predicting a high resolution density map in one go is a challenging task. Hence, we present a two branch CNN architecture for generating high resolution density maps, where the first branch generates a low resolution density map, and the second branch incorporates the low resolution prediction and feature maps from the first branch to generate a high resolution density map. We also propose a multi-stage extension of our approach where each stage in the pipeline utilizes the predictions from all the previous stages.  Empirical comparison with the previous state-of-the-art crowd counting methods shows that our method achieves the lowest mean absolute error on three challenging crowd counting benchmarks: Shanghaitech, WorldExpo'10, and UCF datasets.
\end{abstract}

\section{Introduction}
Gathering of large crowds is commonplace nowadays, and estimating the size of a crowd is an important problem for different purposes ranging from journalism to public safety.  Without turnstiles to provide a precise count, the media and crowd safety specialists must estimate the size of the crowd based on images and videos of the crowd. Manual visual estimation, however, is difficult and laborious for humans. Humans are good at subitizing, i.e., predicting fast and accurate counts for small number of items, but the accuracy with which humans count deteriorates as the number of items increase~\cite{kaufman1949discrimination}. Furthermore, the addition of each new item beyond a few adds an extra processing time of around 250 to 300 milliseconds~\cite{trick1994small}. As a result, any crowd monitoring system that relies on humans for counting people in crowded scenes will be slow and unreliable. There is a need for an automatic computer vision algorithm that can accurately count the number of people in crowded scenes based on images and videos of the crowds. 

There exist a number of computer vision algorithms for crowd counting, and the current state-of-the-art methods are based on \textsl{density estimation} rather than \textsl{detection-then-counting}. Density-estimation methods use Convolutional Neural Networks (CNNs)~\cite{LeCun-et-al-NC89,Krizhevsky-et-al-NIPS12} to output a map of density values, one for each pixel of the input image. The final count estimate can be obtained by summing over the predicted density map. Unlike the detection-then-counting approach (e.g.,~\cite{Hoai-Zisserman-CVPR14}), the output of the density estimation approach at each pixel is not necessarily binary. Density estimation has been proved to be more robust than the detection-then-counting approach because the former does not have to commit to binarized decisions at an early stage. 

\begin{figure}[t]
\centering
\includegraphics[height=0.3\linewidth]{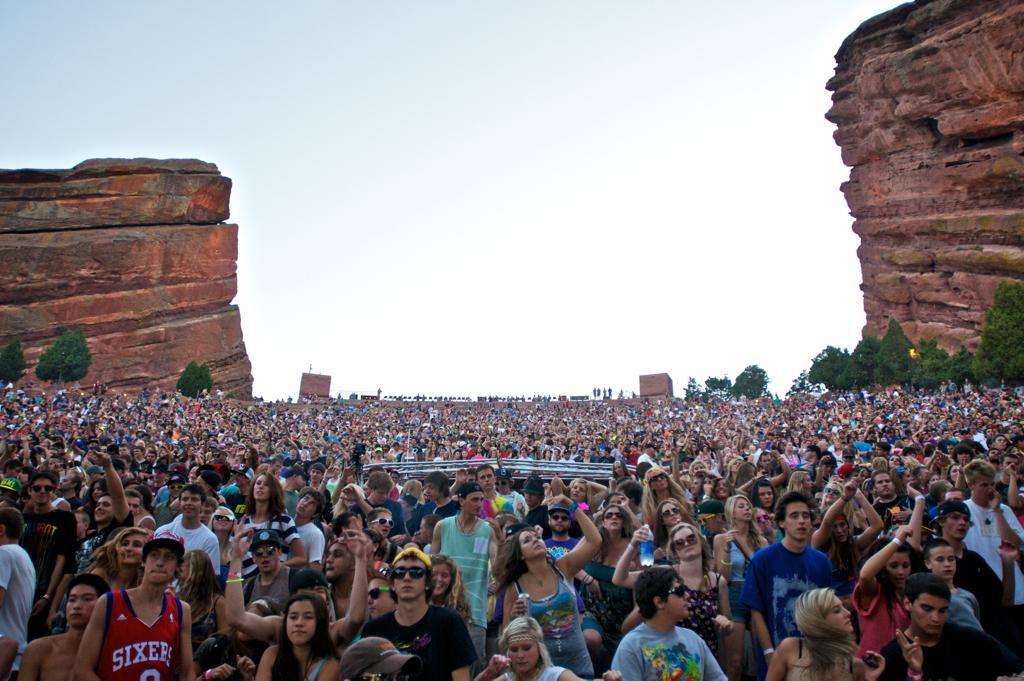}
\includegraphics[height=0.3\linewidth]{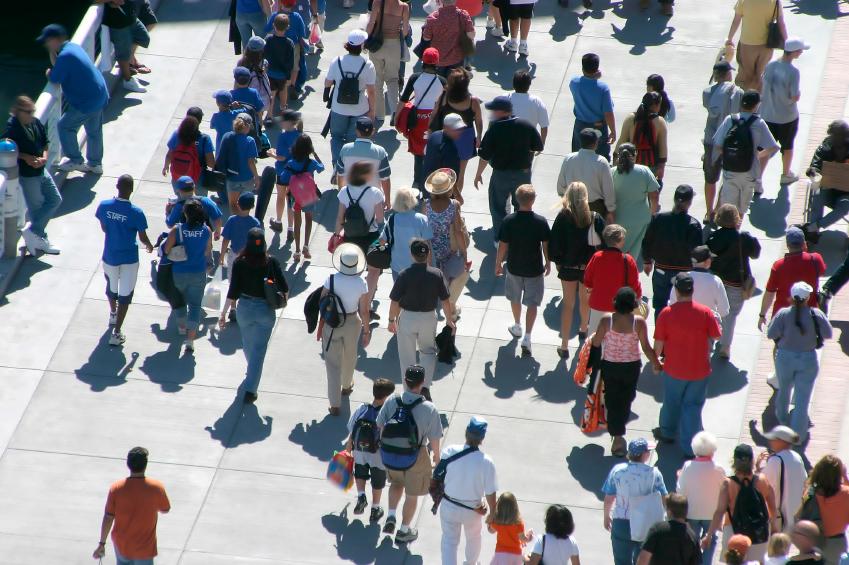}
	\caption{Crowd counting can be posed as a CNN-based density estimation problem, but this problem can be challenging for a single CNN due to the huge variation of density values across pixels of different images. This figure shows two images from the Shanghaitech dataset that have very different crowd densities. As can be seen, crowd count could vary from a few to a few thousand. \label{fig:difdensity}}	
\end{figure}

Estimating the crowd density per pixel is a challenging task due to the large variation of the crowd density values. As shown in Figure~\ref{fig:difdensity}, some images contain hundreds of people, while others have only a few. It is difficult for a single CNN to handle the entire spectrum of crowd densities. Earlier works~\cite{zhang2016single,sam2017switching} have tackled this challenge by using a multi-column or a switching CNN architecture. These CNN architectures consist of three parallel CNN branches with different receptive field sizes. In such  architectures, a branch with smaller receptive fields could handle the high density images well, while a branch with larger receptive fields could handle the low density images. More recently, a five-branch CNN architecture was proposed~\cite{sindagi2017generating} where three of the branches resembled the previous multi-column CNN~\cite{zhang2016single}, while the remaining two branches acted as global and local context estimators. These context estimator branches were trained beforehand on the related task of classifying the image into different density categories. Some of the key takeaways from these previous approaches are: (1) using a multi-column CNN model with varying kernel sizes improves the performance of crowd density estimation; and (2) augmenting the feature set with the ones learned from a task related to density estimation, such as count range classification, improves the performance of the density estimation task.

\begin{figure*}[t]
\centering
\includegraphics[width=\linewidth]{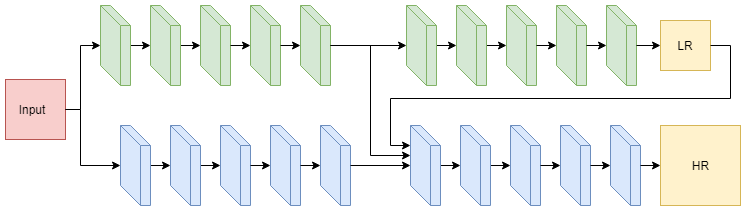}
	\caption{
     Figure shows the ic-CNN architecture which consists of two columns/branches. On the top is the Low Resolution CNN branch (LR-CNN) and at the bottom is the High Resolution CNN branch (HR-CNN). LR-CNN predicts a density map at a lower resolution (LR). It passes the predicted density map and the convolutional feature maps to HR-CNN. HR-CNN fuses its feature maps with the feature maps and predicted density map from LR-CNN, and predicts a high resolution density map (HR) at the size of the original image. LR and HR are low and high resolution prediction maps respectively.
     \label{fig:icCNN}}	
\end{figure*}

In this work, we propose iterative counting Convolutional Neural Networks~(ic-CNN), a CNN-based iterative approach for crowd counting. Unlike previous approaches, where  three~\cite{zhang2016single,sam2017switching} or more~\cite{sindagi2017generating} columns are needed to achieve good performance, our ic-CNN approach has a simpler architecture comprising of two columns/branches. The first branch predicts a  density map at a lower resolution of $\frac{1}{4}$ the size of the original image, and passes the predicted map and a set of convolutional features to the second branch. The second branch predicts a high resolution density map at the size of the original image. Density maps contain information about the spatial distribution of crowd in an image. Hence, the first stage map serves as an important feature for the high resolution density map prediction task. We also propose a multi-stage extension of ic-CNN where we combine multiple ic-CNNs sequentially to further improve the quality of the predicted density map. Each ic-CNN in the multi-stage pipeline provides both the low and high resolution density predictions to all subsequent stages. Figure~\ref{fig:icCNN} illustrates the schematic architecture for ic-CNN. ic-CNN has two branches: Low Resolution CNN (LR-CNN) and High Resolution CNN (HR-CNN). LR-CNN predicts the density map at a low resolution while HR-CNN predicts the density map at the original image resolution. The key highlights of our work are:
 \begin{enumerate}
 \item We propose ic-CNN, a two-stage CNN framework for crowd density estimation and counting.
 
 \item ic-CNN achieves state of the art results on multiple crowd counting datasets. On Shanghaitech Part B dataset, ic-CNN yields $48.3\%$ improvement in terms of mean absolute error over the previously published results~\cite{sindagi2017generating}. 
 \item We also propose a multi-stage extension of ic-CNN, which can combine
 predictions from multiple ic-CNN models. 
 \end{enumerate}

\section{Related Work} \label{sec:RelatedWork}
Crowd counting is an important research problem and a number of approaches have been proposed by the computer vision community. Earlier work tackled crowd counting as an object detection problem~\cite{li2008estimating,lin2001estimation}. Lin \textit{et al.}~\cite{lin2001estimation} extracted Haar features for head like contours and used an SVM classifier to classify these features as the contour of a head or not. Li \textit{et al.} \cite{li2008estimating} proposed a detection based approach where the input image was first segmented into foreground-background regions and a HOG feature based head-shoulder detector was used to detect each person in the crowd. These detection based methods often fail to accurately count people in extremely dense scenes. To handle images of dense crowds, some methods~\cite{chan2009bayesian,chen2012feature} proposed to use a regression approach to avoid the harder detection problem. They instead extracted local patch level features and learned a regression function to directly estimate the total count for an input image patch. These regression approaches, however, do not fully utilize the available annotation associated with training data; they ignore the spatial density and distribution of people in training images. Several researchers~\cite{lempitsky2010learning,pham2015count} proposed to use a density estimation approach to take advantage of the provided crowd density annotation maps of training images. Lempitsky \& Zisserman~\cite{lempitsky2010learning} learned a linear mapping between the crowd images and the corresponding ground truth density maps. Pham~\textit{et al.}~\cite{pham2015count} learned a more robust mapping by using a random decision forest to estimate the crowd density map. These density-based methods solve some of the challenges faced by the earlier detection and regression based approaches, by avoiding the harder detection problem and also utilizing the spatial annotation and correlation. All aforementioned methods predated the deep-learning era, and they used hand crafted features for crowd counting. 


More recent methods~\cite{wang2015deep,fu2015fast,zhang2016single,sam2017switching,onoro2016towards,sindagi2017generating} used CNNs to tackle crowd counting. Wang \textit{et al.}~\cite{wang2015deep} posed crowd counting as a regression problem, and used a CNN model to map the input crowd image to its corresponding count. Instead of predicting the overall count, Fu \textit{et al.}~\cite{fu2015fast} classified an image into five broad crowd density categories and used a cascade of two CNNs in a boosting like strategy where the second CNN was trained on the images misclassified by the first CNN. These methods also overlooked the benefits provided by the crowd density annotation maps. 

The methods that are most related to our work are \cite{zhang2016single,sam2017switching,sindagi2017generating}. Zhang~\textit{et al.}~\cite{zhang2016single} proposed a CNN-based method to predict crowd density maps. To handle the large variation in crowd densities and sizes across different images,  Zhang~\textit{et al.}~\cite{zhang2016single} proposed a multi-column CNN architecture (MCNN) with filters and receptive fields of various sizes. The CNN column with smaller receptive field and filter sizes were responsible for the denser crowd images, while the CNN columns with larger receptive fields and filter sizes were meant for the less dense crowd images. The features from the three columns were concatenated and processed by a \mbox{$1$$\times$$1$} convolution layer to predict the final density map. To handle the variations in density and size within an image, the authors divided each image into non-overlapping patches, and trained the MCNN architecture on these patches. 
Given that the number of training samples in annotated crowd counting datasets is much smaller in comparison to the datasets pertaining to image classification and segmentation tasks, training a CNN from scratch on full images might lead to overfitting. Hence, patch-based training of MCNN was essential in preventing overfitting and also improving the overall  performance by serving as a data augmentation strategy. One issue with MCNN was that it fused the features form three CNN columns for predicting the density map. For a given patch, it is expected that the counting performance can be made more accurate by choosing the right CNN column that specializes in analyzing images of similar density values. 
Sam \textit{et al.}~\cite{sam2017switching} built on this idea and decoupled the three columns into separate CNNs, each focused on a subset of the training patches. To decide which CNN to assign a patch to, the authors trained a CNN-based switch classifier. However, since the ground truth label needed to train the switch classifier was unavailable, the authors resorted to a multi-stage training strategy: 1) training the three density predicting CNNs on the entire set of training patches, 2) training the switch classifier using the count from the previous stage to decide the switch labels, and 3) retraining the three CNNs using the patches assigned by the switch classifier.
In a more recent work, Sindagi \textit{et al.}~\cite{sindagi2017generating} further modified the MCNN architecture by adding two more branches for estimating global and local context maps. The global/local context prediction branches were trained beforehand for the related task of classifying an image/patch into five different count categories. The classification scores were used to create a feature map of the same size as the image/patch, which served as the global/local context map. These context maps were fused with the convolutional feature maps obtained using a three branch multi-column CNN, and the resulting features were further processed by convolutional layers and a $1$$\times$$1$ convolution layer to obtain the final density map.
\section{Proposed Approach}
In this section, we describe the architecture of ic-CNN, its multi-stage extension
and the training strategy.
ic-CNN is discussed in Section \ref{sec:BasicCountingBlock}. 
The multi stage extension of ic-CNN is discussed in Section~\ref{sec:ic-CNN}, and the training details are discussed in Sec~\ref{sec:TrainingDetails}.

\subsection{Iterative Counting CNN}\label{sec:BasicCountingBlock}

Let $\mD = \{(X_1,Y_1,Z_1),\dotsc,(X_n,Y_n,Z_n)\}$ be the training set of $n$ (image, high resolution density map, low resolution density map) triplets, where $X_i$ is the $i^{th}$ image, $Y_i$ is the corresponding crowd density map at the same resolution as the image $X_i$, and $Z_i$ is a low resolution version of the crowd density map. $Y_i$ and $Z_i$ have the same overall count. Let $f_l$ and $f_h$ be the mapping functions which transform the image into the low resolution and high resolution density maps, respectively. Let the parameters of the low resolution branch (LR-CNN) and high resolution branch (HR-CNN) be $\theta_l$ and $\theta_h$ respectively. Note that $f_l$ depends on only $\theta_l$, while $f_h$ depends on both $\theta_l$ and $\theta_h$. Given an input image $X_i$, the low resolution density map $\hat{Z_i}$ can be obtained by a doing a forward pass through the LR-CNN branch:
\begin{equation}\label{eqn:op1}
\hat{Z_i}= f_l(X_i;\theta_l). 
\end{equation}
The inputs to the high resolution branch HR-CNN are: the image $X_i$, the features computed by the low resolution branch LR-CNN, and the low resolution prediction $\hat{Z_i}$. HR-CNN predicts a high resolution density map of the same size as the original image:
\begin{equation}\label{eqn:op2}
\hat{Y_i}= f_h(X_i,\hat{Z_i};\theta_l, \theta_h). 
\end{equation}

The low resolution prediction $\hat{Z_i}$ contains information about the spatial distribution of the crowd in the image $X_i$. It serves as an important feature map for the high resolution prediction task. We can learn the parameters $\theta_l$  and $\theta_h$ by minimizing
the loss function $\mL(\theta_l, \theta_h)$:
\begin{equation}\label{eqn:loss1}
\mL(\theta_l, \theta_h) = \frac{1}{n} \sum_{i=1}^{n}( \lambda_{l}L(f_l(X_i;\theta_l),Z_i) +  \lambda_{h} L(f_h(X_i,\hat{Z}_i;\theta_l, \theta_h),Y_i)), 
\end{equation}
where $L(\cdot, \cdot)$ denotes the loss function, and a reasonable choice is to use the squared error between the estimated and ground truth values. $ \lambda_{l}$ and  $\lambda_{h}$ are scalar hyperparameters which can be used to give more importance to one of the loss terms.
Using Equations (\ref{eqn:op1}) and (\ref{eqn:op2}), the right hand side can be further simplified as:
\begin{equation}\label{eqn:loss2}
\mL(\theta_l, \theta_h) = \frac{1}{n} \sum_{i=1}^{n}(\lambda_l L(\hat{Z}_i,Z_i) +  \lambda_h L(\hat{Y}_i,Y_i)).
\end{equation}

At test time, given an image $X_i$, we first obtain the low resolution output $\hat{Z_i}$ by doing a forward pass through LR-CNN and then pass the convolutional features and the low resolution map $\hat{Z_i}$ to HR-CNN, which will predict the high resolution map  $\hat{Y_i}$. We use the high resolution output predicted by HR-CNN as the final output of ic-CNN. The overall crowd count is obtained by summing over all the pixels in the density map $\hat{Y_i}$.\
\newline Below we provide the architecture details for the LR-CNN and HR-CNN branches.
\newline
\myheading{LR-CNN.} The LR-CNN branch takes as input an image, and predicts a density map at $\frac{1}{4}$ the size of the original image. LR-CNN has the following architecture:
Conv3-64, Conv3-64, MaxPool, Conv3-128, Conv3-128, MaxPool, Conv3-256, Conv3-256, Conv3-256, Conv7-196, Conv5-96, Conv3-32, Conv1-1. Here, ConvX-Y implies a convolution layer having Y filters with \htimesw{X}{X} kernel size. MaxPool is the max pooling layer. We use a ReLU nonlinearity after each convolutional layer.

\myheading{HR-CNN.}
The HR-CNN branch predicts the high resolution density map at the same size as the input image. HR-CNN has the following architecture:
Conv7-16, MaxPool, Conv5-24, MaxPool, Conv3-48, Conv3-48, Conv3-24, Conv7-196, Conv5-96, Upsampling-2, Conv3-32, Upsampling-2, Conv1-1. Here, Upsampling-2 is a bilinear interpolation layer which upsamples the input to twice its size.
\subsection{Multi-stage Crowd Counting}\label{sec:ic-CNN}
A multi-stage ic-CNN is a network that combines multiple building blocks of ic-CNN described in the previous section. Each ic-CNN block inputs the low and high resolution prediction maps from all the previous blocks. Given an input image $X_i$, the low resolution branch of the $k^{th}$ block, represented by the function $f_{l}^{k}$, outputs the low resolution prediction:
\begin{equation}\label{eqn:op1a}
\hat{Z}_{i}^k= f_{l}^k(X_i,\hat{Z}_{i}^{1:k-1},\hat{Y}_{i}^{1:k-1},\theta_{l}^k), 
\end{equation}
where $\theta_{l}^k$ represents the parameters of LR-CNN, $\hat{Z}_{i}^{1:k-1}$ and $\hat{Y}_{i}^{1:k-1}$ represent the set of low and high level predictions from the first $k-1$ blocks for the input $X_i$.
The high resolution branch of the $k^{th}$ block, represented by the function $f_{h}^{k}$, takes as input the image $X_i$, the feature maps computed by the low resolution branch $f_{l}^k$, the low resolution prediction $\hat{Z}_{i}^k$, and the entire set of low and high resolution prediction maps from the first $k-1$ blocks. Hence, the output of the $k^{th}$ HR-CNN can be computed using: 
\begin{equation}\label{eqn:op2a}
\hat{Y}_{i}^{k}= f_{h}^{k}(X_i,\hat{Z}_{i}^{1:k},\hat{Y}_{i}^{1:k-1},\theta_{l}^k, \theta_{h}^k).
\end{equation}
Note that $f_{l}^k$ and $f_{h}^k$ do not depend on the parameters for the first $k-1$ blocks, and $\hat{Z}_{i}^{1:k-1}$ and $\hat{Y}_{i}^{1:k-1}$  are treated as fixed inputs (i.e., the parameters of the corresponding network blocks are frozen). We can learn the parameters $\theta_{l}^k$  and $\theta_{h}^k$ by minimizing the loss function $\mL(\theta_{l}^k, \theta_{h}^k)$:
\begin{align}\label{eqn:loss2}
\mL(\theta_{l}^k, \theta_{h}^k) = &\frac{\lambda_{l}}{n} \sum_{i=1}^{n} L(f_{l}^k(X_i,\hat{Z}_{i}^{1:k-1},\hat{Y}_{i}^{1:k-1},\theta_{l}^k),Z_i) \nonumber \\
 + &  \frac{\lambda_{h}}{n} \sum_{i=1}^{n} L(f_{h}^{k}(X_i,\hat{Z}_{i}^{1:k},\hat{Y}_{i}^{1:k-1},\theta_{l}^k, \theta_{h}^k),Y_i).
\end{align}

\subsection{Training Details}\label{sec:TrainingDetails}
An ic-CNN is trained by minimizing the loss function $\mL(\theta_l,\theta_h)$ from Equation~(\ref{eqn:loss1}). We use the Stochastic Gradient Descent algorithm with the following hyper parameters (unless specified otherwise): learning rate $10^{-4}$, momentum $0.9$, batch size 1. We give more importance to the high resolution loss term in Equation~(\ref{eqn:loss1}) and set $\lambda_{l}$ and $\lambda_{h}$ to $10^{-2}$ and $10^{2}$, respectively.

We train a multi-stage ic-CNN in multiple stages. In the $k^{th}$ stage, we train the $k^{th}$ ic-CNN block by minimizing the loss function given in Equation~(\ref{eqn:loss2}), using the Stochastic Gradient Descent algorithm with the same hyper parameters as above. Once the training for the $k^{th}$ stage has converged, we freeze the parameters for the $k^{th}$ stage and proceed to the next stage. 

The training data consists of crowd images and corresponding ground truth annotation files. A ground truth annotation for an image specifies the location of each person in the image with a single dot on the person.  We convert this annotation into a binary map consisting of $0$'s at all locations, except for the annotated points which are assigned the value  of $1$. We convolve this binary map with a Gaussian filter of standard deviation $5$. We use the resulting density map for training the networks.

\section{Experiments}
We conduct experiments on three challenging datasets: Shanghaitech~\cite{zhang2016single}, WorldExpo'10~\cite{zhang2015cross}, and UCF Crowd Counting Dataset~\cite{idrees2013multi}.

\subsection{Evaluation Metrics}
Following previous works for crowd counting, we use the Mean Absolute Error (MAE) and Root Mean Squared Error (RMSE) to evaluate the performance of our proposed method. If the predicted count for image $i$ is $\hat{C_i}$ and
the ground truth count is $C_i$, the MAE and RMSE can be computed as: 
\begin{align}
MAE = \frac{1}{n}\sum_{i=1}^{n} \lvert C_i - \hat{C_i} \rvert, \hspace{3ex} RMSE = \sqrt[]{\frac{1}{n}\sum_{i=1}^{n} (C_i - \hat{C_i})^2}
\end{align}
where $n$ is the number of test images.

\subsection{Experiments on the Shanghaitech Dataset}

\setlength{\tabcolsep}{14pt}
\begin{table}[t]
\begin{center}	
\caption{{\bf Count errors of different methods on the Shanghaitech dataset.}  This dataset has two parts: A and B. We compare  ic-CNN with the previous state-of-the-art approaches, using two metrics: Mean Absolute Error (MAE) and Root Mean Squared Error (RMSE). ic-CNN (one stage) is the single stage ic-CNN with two branches HR-CNN and LR-CNN. ic-CNN (two stages) is the two-stage variant of ic-CNN. Both ic-CNN networks outperform the previous approaches in 3 out of 4 cases. On the Shanghaitech Part B dataset, using the one-stage ic-CNN, which has a simpler architecture than CP-CNN~\cite{sindagi2017generating}, we improve on the previously reported state of the art results by  $48.3\%$ using the MAE metric and  $46.8\%$ using the RMSE metric  \label{table_shanghaitech}
}
\begin{tabular}{lrrrr}
\toprule
         & \multicolumn{2}{c}{Part A} & \multicolumn{2}{c}{Part B} \\
         \cmidrule(lr){2-3} \cmidrule(lr){4-5} 
               & MAE          & RMSE         & MAE          & RMSE         \\
\midrule
Crowd CNN~\cite{zhang2015cross}     &        181.8      &  277.7           &   32.0           &  49.8           \\


MCNN~\cite{zhang2016single}           &         110.2     &    173.2         &  26.4            &  41.3           \\
Switching CNN~\cite{sam2017switching}  &   90.4           &  135.0           &     21.6         &     33.4        \\
CP-CNN~\cite{sindagi2017generating}  & 73.6 &\textbf{106.4} &20.1 & 30.1  \\
ic-CNN (one stage) & 69.8&117.3& \textbf{10.4}& 16.7\\
ic-CNN (two stages)  &\textbf{68.5}&116.2& \textbf{10.7}& \textbf{16.0}\\
\bottomrule
\end{tabular}
\end{center}
\end{table}

The Shanghaitech dataset~\cite{zhang2016single} consists of
$1198$ annotated crowd images. The dataset is divided into two parts, Part-A containing $482$ images and Part-B containing $716$ images. Part-A is split into train and test subsets consisting of 300 and 182 images,  respectively. Part-B is split into train and test subsets consisting of 400 and 316 images. Each person in a crowd image is annotated with one point close to the center of the head. In total, the dataset consists of 330,165 annotated people. Images from Part-A were collected from the Internet, while images from Part-B were collected on the busy streets of Shanghai. To avoid the risk of overfitting to the small number of training images, we trained ic-CNNs on random crops of size $\frac{H}{3}\times\frac{W}{3}$, where $H$ and $W$ are the height and width of a training image. In Table~\ref{table_shanghaitech}, we compare ic-CNNs with the previous state-of-the-art approaches. ic-CNNs outperform the previous approaches in three out of four cases by a large margin. On Part-B of the Shanghaitech dataset, using the one-stage ic-CNN which has a simpler architecture than the five-branch CP-CNN~\cite{sindagi2017generating}, we improve on the previously reported state of the art results by $48.3\%$ for MAE metric and $46.8\%$ for the RMSE metric. On Part A of the Shanghaitech dataset, we achieve a $5.1$ absolute improvement in MAE over CP-CNN. Furthermore, for Part A data, the two-stage ic-CNN results in an improvement of $1.3$ MAE over the one-stage ic-CNN. We also trained a three-stage ic-CNN on Part A data,  which resulted in MAE = 69.4 and RMSE = 116.0. Since adding the $3^{rd}$ stage did not yield a significant performance gain, we did not experiment with more than three stages.

In Table~\ref{tab:Rebuttal-resolution}, we analyze the effects of varying the resolution of the intermediate prediction on the overall performance. Using any resolution other than $\frac{1}{4}$ leads to a drop in the performance.

\begin{table}[!t]
\centering
   \caption{MAE and RMSE on Shanghaitech Part-A dataset as we vary the resolution being used for the low resolution branch LR-CNN of ic-CNN. The resolution of HR-CNN is fixed at 1, the size of the input image.\label{tab:Rebuttal-resolution} }
\begin{tabular}{cccc}
\toprule
LR-Resolution & HR-Resolution & MAE & RMSE \\
\midrule
1/8   &  1  &  74.9   &    131.6 \\ 
1/4  & 1   &  \textbf{69.8}   &   \textbf{117.3}  \\  
1/2  & 1   &  73.3  &  124.4  \\ 
1   &  1  &   74.4  &  128.3   \\    \bottomrule
   \end{tabular}
\end{table}

\setlength{\tabcolsep}{25pt}
\begin{table}[!t]
\centering
\caption{{\bf Effect of varying hyper parameter $\lambda_{h}$:} Mean absolute error on Shanghaitech Part A dataset. $\lambda_{l}$ is kept fixed at $10^{-2}$.
}
\begin{tabular}{lcc}
\toprule
$\lambda_{h}$     & LR-CNN   & HR-CNN   \\
\midrule
$10^{-4}$  & 73.7  & 78.8  \\
$10^{-2}$  & 73.0  & 73.6 \\
$1$   & 75.1  & 73.3 \\
$10^{2}$  & 79.9  &  69.8 \\
$10^{4}$  &  432.6  & 74.4 \\
\bottomrule
\end{tabular}
\label{tab:tableLossWeights}
\end{table}

 In Table~\ref{tab:tableLossWeights}, we analyze the effects of varying the hyperparameter $\lambda_{h}$ on performance of ic-CNN. We use Shanghaitech Part-A dataset for this experiment. We show the MAE of the high and low resolution branches as the scalar weight $\lambda_{h}$ is varied. $\lambda_{l}$ is kept fixed at $10^{-2}$. We can see that the
LR-CNN branch performs better when $\lambda_{l}$ is comparable with $\lambda_{h}$, and its performance degrades when $\lambda_{h}$ is too large. The performance of HR-CNN improves as $\lambda_{h}$ is varied from $10^{-4}$ to $10^2$. In the extreme case when $\lambda_{h}$ is set to $10^{4}$, there is a large degradation in the performance of the LR-CNN branch, which affects the performance of the HR-CNN branch. When $\lambda_{h}$ is $10^{4}$, the low resolution prediction task is possibly ignored, and the network solely focuses on solving the high resolution task. In such a scenario, the low resolution prediction does not contain any useful information, which affects the performance of the high resolution branch HR-CNN. We obtain the best results for the HR-CNN branch
when $\lambda_{h}$ is set to $10^{2}$. In this case, the high resolution loss does not force the network to completely ignore the low resolution task.

In Table~\ref{tab:Rebuttal-trainingtime}, we show the training time and the number of parameters of ic-CNN, MCNN, Switching CNN, and CP-CNN. An ic-CNN takes 10 hours to train, while a Switching CNN takes around 22 hours. An ic-CNN has significantly fewer parameters than a CP-CNN and a Switching CNN. We contacted the authors of MCNN and CP-CNN, but we did not get a response for the training time of these networks.

\setlength{\tabcolsep}{10pt}
\begin{table}[!t]
\centering
\caption{Training time, number of parameters, and MAE on Part A of the Shanghaitech dataset. ic-CNN was trained on a single GPU  machine (Nvidia GTX 1080 TI). \label{tab:Rebuttal-trainingtime}}
\begin{tabular}{lccc}
\toprule
Model         & Training Time  &Number of Parameters & MAE \\
\midrule

MCNN ~\cite{zhang2016single}         &     unknown          &         $1.27 \times 10^5$        & 110.2    \\
Switching CNN~\cite{sam2017switching} &     22 hrs                 &    $1.2 \times 10^7$      &  90.4   \\
CP-CNN ~\cite{sindagi2017generating}       & unknown         &         $6.3 \times 10^7$         & 73.6    \\
ic-CNN (proposed)  &     10 hrs                  &    $7.9 \times 10^6$       & 69.8 \\
\bottomrule
\end{tabular}
\end{table}

In Table~\ref{tab:tableablation}, we analyze the importance of each of the components of our proposed ic-CNN model. We see that both the feature sharing  and the feedback of the low resolution prediction are important for ic-CNN. Removing any of these two components leads to significant drop in performance.

\begin{table}[!t]
\centering
\caption{{\bf Ablation study on Shanghaitech Part A data.} HR-CNN is the high resolution branch, LR-CNN is the low resolution branch. LR-CNN alone and HR-CNN alone  refer to a counting network that contains either LR-CNN or HR-CNN only. ic-CNN is our proposed approach, where both the features and the low resolution prediction map from LR-CNN are shared with HR-CNN. We also compared with two variants where either the low resolution map or the convolutional feature maps from LR-CNN is not shared with the HR-CNN. \label{tab:tableablation}}
\begin{tabular}{lrr}
\toprule
Method     & MAE   & RMSE   \\
\midrule
LR-CNN alone   & 78.5 & 133.2\\
HR-CNN alone  & 136.2  & 204.0 \\
HR-CNN + LR-CNN features (no low-res prediction) & 75.1  & 129.0\\
HR-CNN + LR-CNN low-res prediction (no features) & 77.4 & 130.4 \\
ic-CNN (proposed)  & \textbf{69.8} &  \textbf{117.3}\\
\bottomrule
\end{tabular}
\end{table}

\begin{figure}[!thb]
\centering
\includegraphics[scale=0.3]{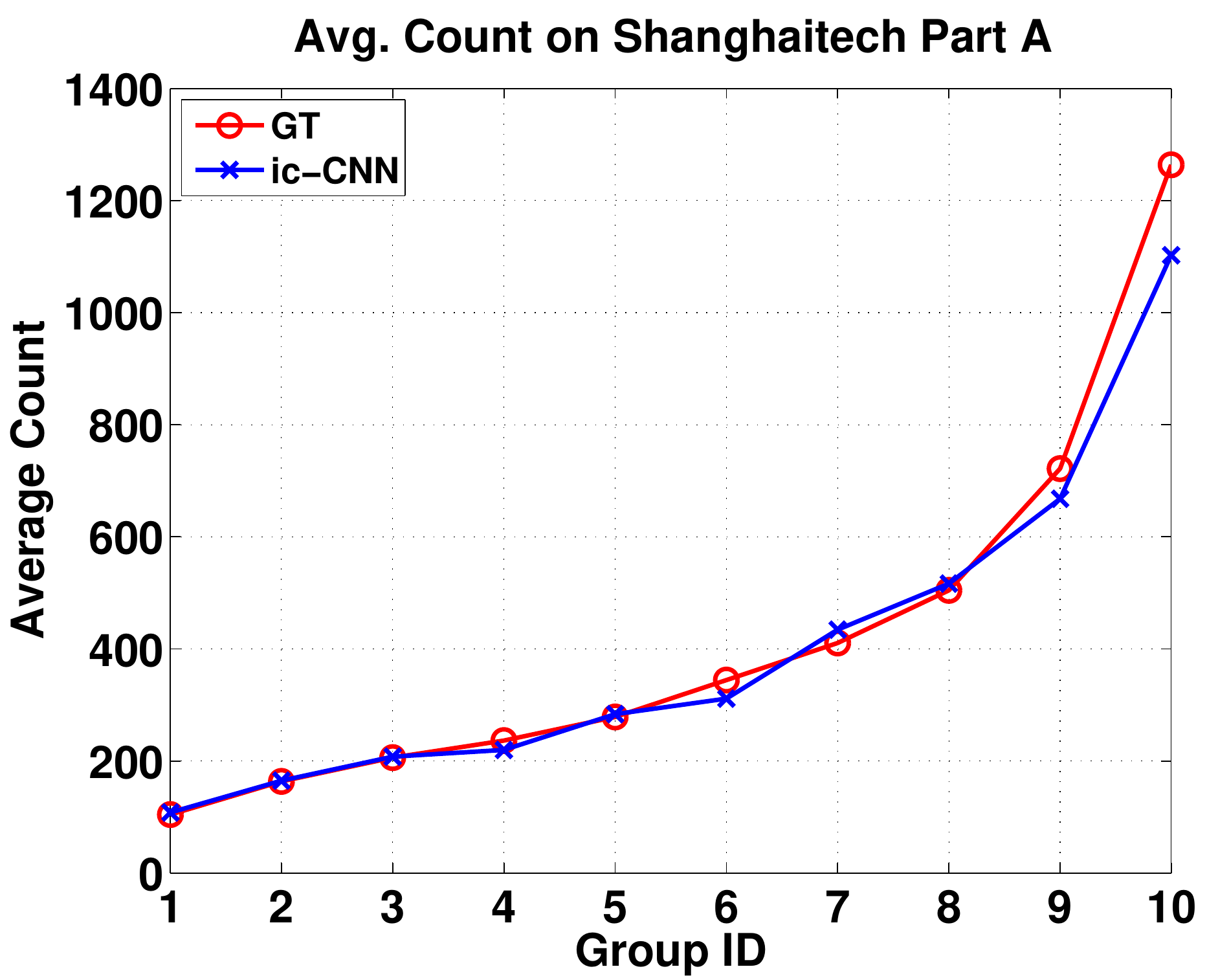}

	\caption{\textbf{Performance across different crowd density:} We divide the 182 test images from Shanghaitech Part A into 10 groups on the basis of the crowd count. Each group except the last has 18 test images. We average the crowd count across a group to obtain the average count. GT is the ground truth, ic-CNN is prediction from the high resolution branch. For majority of the count groups, the difference between the average counts for ic-CNN and GT is small.\label{fig:avgcount}}	
\end{figure}
In Figure~\ref{fig:avgcount}, we analyze the performance of ic-CNN across different groups of images with varying crowd counts.
\subsection{Experiments on the WorldExpo'10 Dataset} 
The WorldExpo'10 dataset consists of $1132$ annotated video sequences captured by $108$ surveillance cameras. Annotated frames from $103$ cameras are used for training and the annotated frames from the remaining $5$ cameras are used for testing. We trained ic-CNN networks using random crops of sizes $\frac{H}{2}\times\frac{W}{2}$. We used the networks trained on Shanghaitech Part A for initializing the models for the experiments on the WorldExpo dataset. In Table \ref{tab:worldexpo10}, we compare ic-CNN with other state of art approaches. ic-CNN outperforms these previous approaches on three out of five cases.

\setlength{\tabcolsep}{7pt}
\begin{table}[!tbp]
\centering
\caption{{\bf Performance of different methods on the WorldExpo'10 dataset}.  Switch CNN(with perspective) refers to the case when perspective maps are used to obtain the crowd density map, while Switch CNN(sans perspective) refers to the case when the perspective map isn't used.ic-CNN is our proposed two branch approach. We outperform other approaches on 3 of 6 cases. \label{tab:worldexpo10}}
\begin{tabular}{lrrrrrr}
\toprule 
Method                         & S1 & S2 & S3 & S4 & S5 & Avg \\
\midrule
Crowd CNN~\cite{zhang2015cross}                   &  9.8  & 14.1   &  14.3  & 22.2   & \textbf{3.7}   &  12.9   \\
MCNN  ~\cite{zhang2016single}                         &  3.4  &  20.6  &  12.9  & 13.0   & 8.1   &   11.6  \\
Switching CNN (sans perspective)~\cite{sam2017switching}  &  4.4  &  15.7  & 10.0   & 11.0   & 5.9   & 9.4 \\
Switching CNN (with perspective)~\cite{sam2017switching} & 4.2   &   14.9 & 14.2   &  18.7  & 4.3   & 11.2 \\
CP-CNN\cite{sindagi2017generating} & \textbf{2.9 }& 14.7 &10.5
& 10.4 & 5.8 & \textbf{8.8 }\\
ic-CNN (proposed) & 17.0 & \textbf{12.3} &  \textbf{9.2}  & \textbf{8.1} & 4.7 & 10.3 \\
\bottomrule
\end{tabular}
\end{table}
 
\setlength{\tabcolsep}{25pt}
\subsection{Experiments on the UCF Dataset}
\begin{table}[!h]
\centering
\caption{{\bf Performance of various methods on the UCF Crowd Counting dataset}. The proposed method ic-CNN achieves the best MAE. \label{tab:tableucf}}
\begin{tabular}{lrr}
\toprule
Method     & MAE   & RMSE   \\
\midrule
Lempitsky \& Zisserman~\cite{lempitsky2010learning}  & 493.4 & 487.1 \\
Idrees et. al~\cite{idrees2013multi}     &  419.5     & 487.1      \\
Crowd CNN ~\cite{zhang2015cross}     &    467.0   & 498.5      \\
Crowdnet~\cite{boominathan2016crowdnet}   &     452.5  & -      \\
MCNN~\cite{zhang2016single}      &   377.6    &  509.1     \\
Hydra2s ~\cite{onoro2016towards}  &    333.7   &  425.6    \\
Switch CNN~\cite{sam2017switching} &     318.1  &  439.2     \\
CP-CNN~\cite{sindagi2017generating} &295.8 & \textbf{320.9} \\
ic-CNN (proposed) &\textbf{260.9} & 365.5 \\
\bottomrule
\end{tabular}
\vskip 0.1in
\end{table}

The UCF Crowd Counting dataset~\cite{idrees2013multi} consists of $50$ crowd images collected from the web. Each person in the dataset is annotated with a single dot annotation. The numbers of people in the images vary from $94$ to $4545$ with an average of $1280$ people per image. The average count for the UCF dataset is much larger than the previous two datasets. Following previous works using this dataset,  we performe five-fold cross validation and report the MAE and RMSE values. We trained ic-CNN networks using random crops of sizes $\frac{H}{3}\times\frac{W}{3}$. We compare ic-CNN with previous approaches and show the results in Table~\ref{tab:tableucf}. Since the dataset is small, adding multiple stages to ic-CNN could lead to overfitting. Hence we only use one-stage ic-CNN on the UCF dataset. ic-CNN achieves the best MAE on this dataset, outperforming CP-CNN by a large margin.

\subsection{Qualitative Results}
\begin{figure}
\begin{tabular}{cccc} 

{\bf Image} & {\bf Ground truth} & {\bf LR output} & {\bf HR  output} \\ \\ 
{  
	\includegraphics[width=0.22\linewidth]{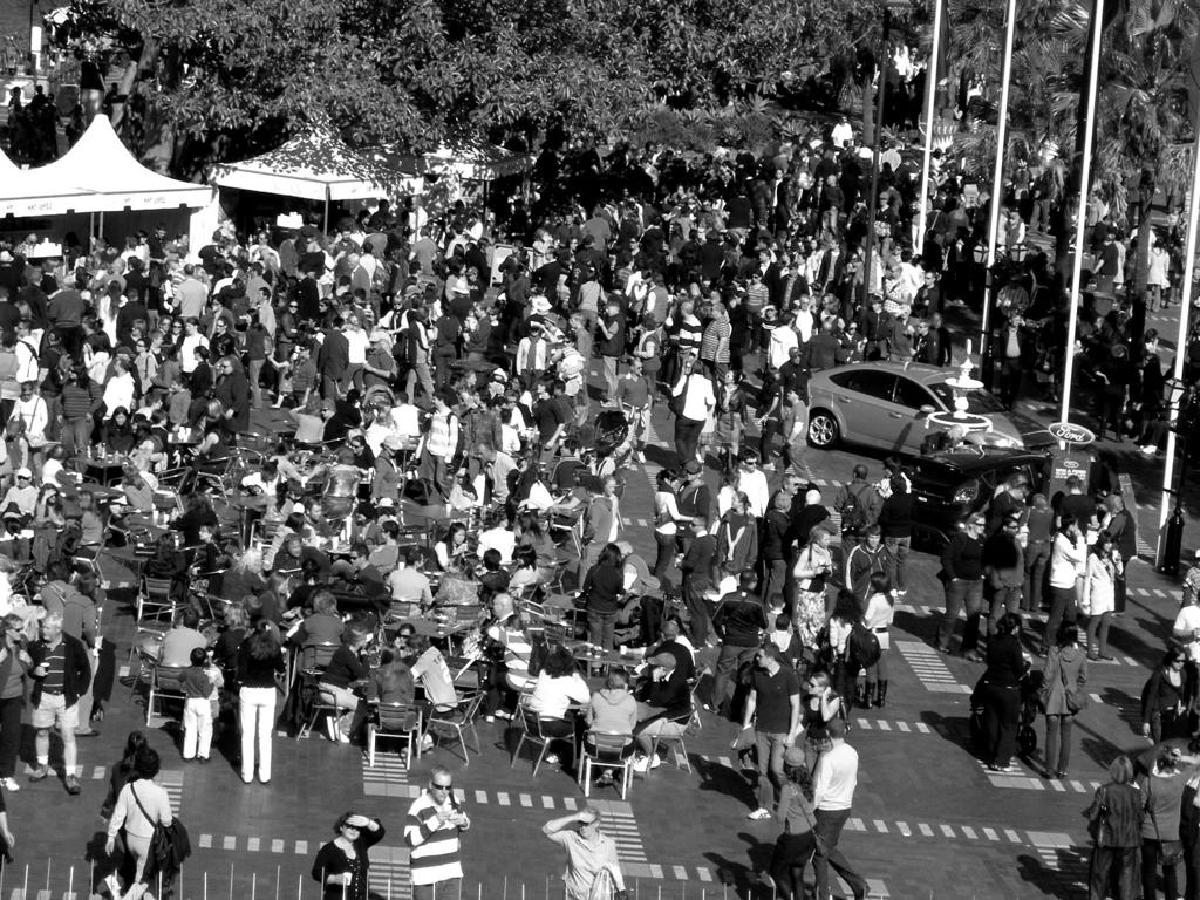}
     } &  {  
	\includegraphics[width=0.22\linewidth]{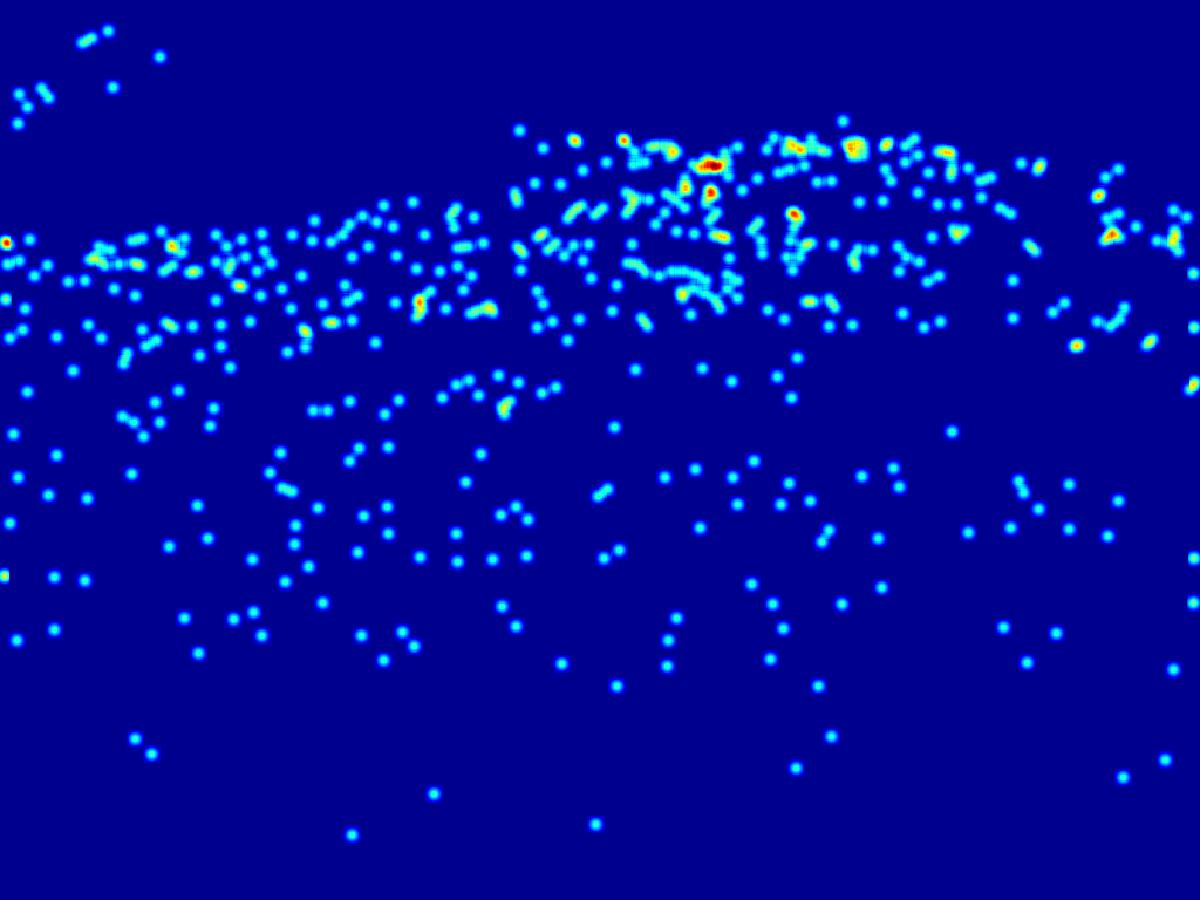}
     }
     &
     {\includegraphics[width=0.22\linewidth]{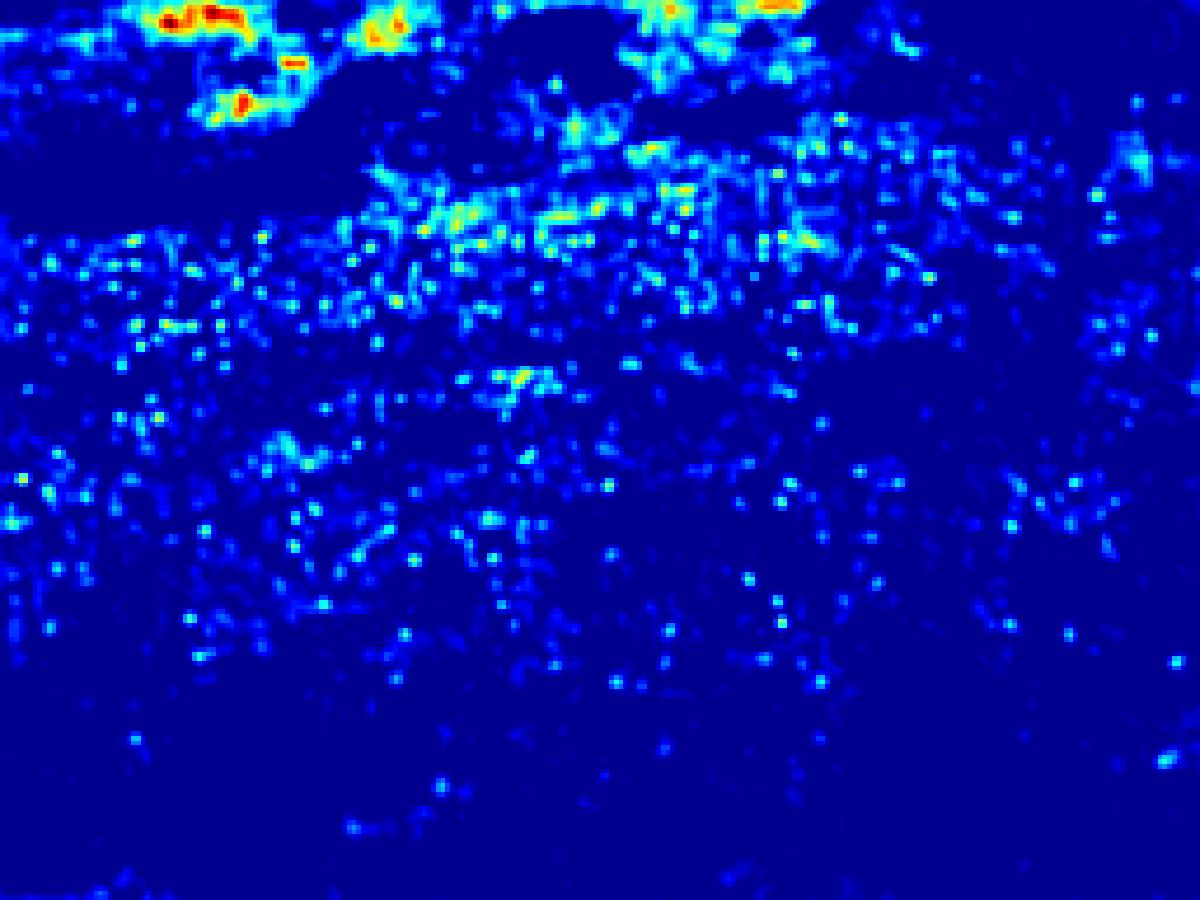}
     } & 
     {\includegraphics[width=0.22\linewidth]{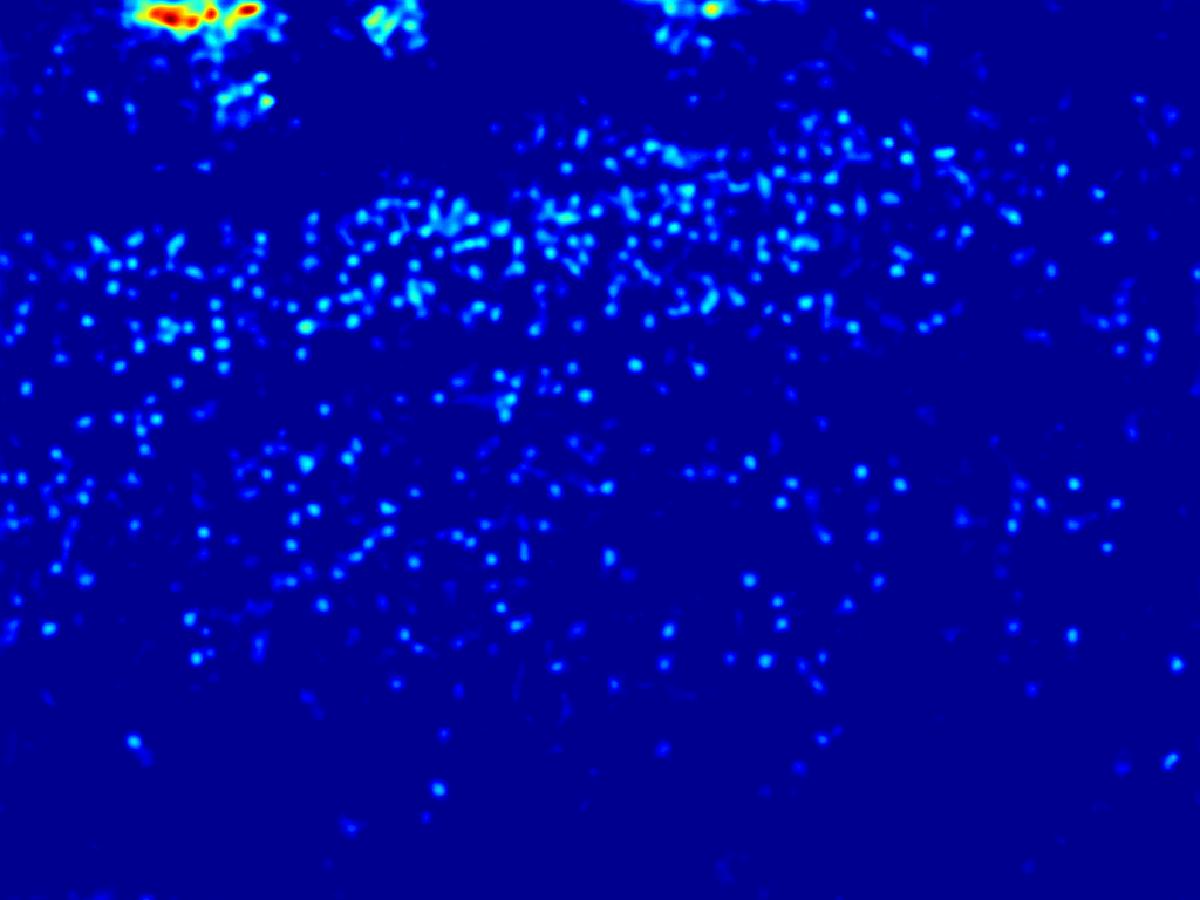}
     } \\
    &502  & 793 & 512\\
   \\ 
{  
	\includegraphics[width=0.22\linewidth]{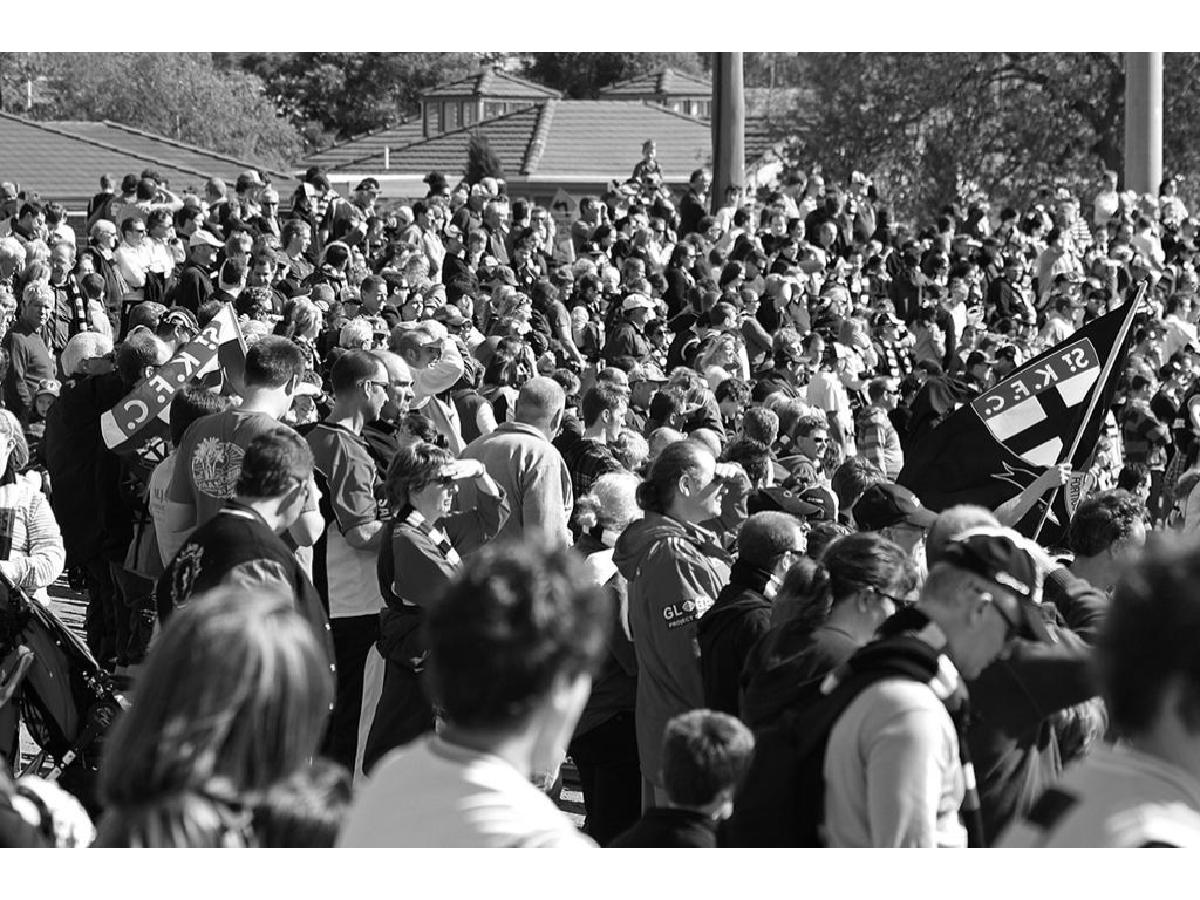}
     } &  {  
	\includegraphics[width=0.22\linewidth]{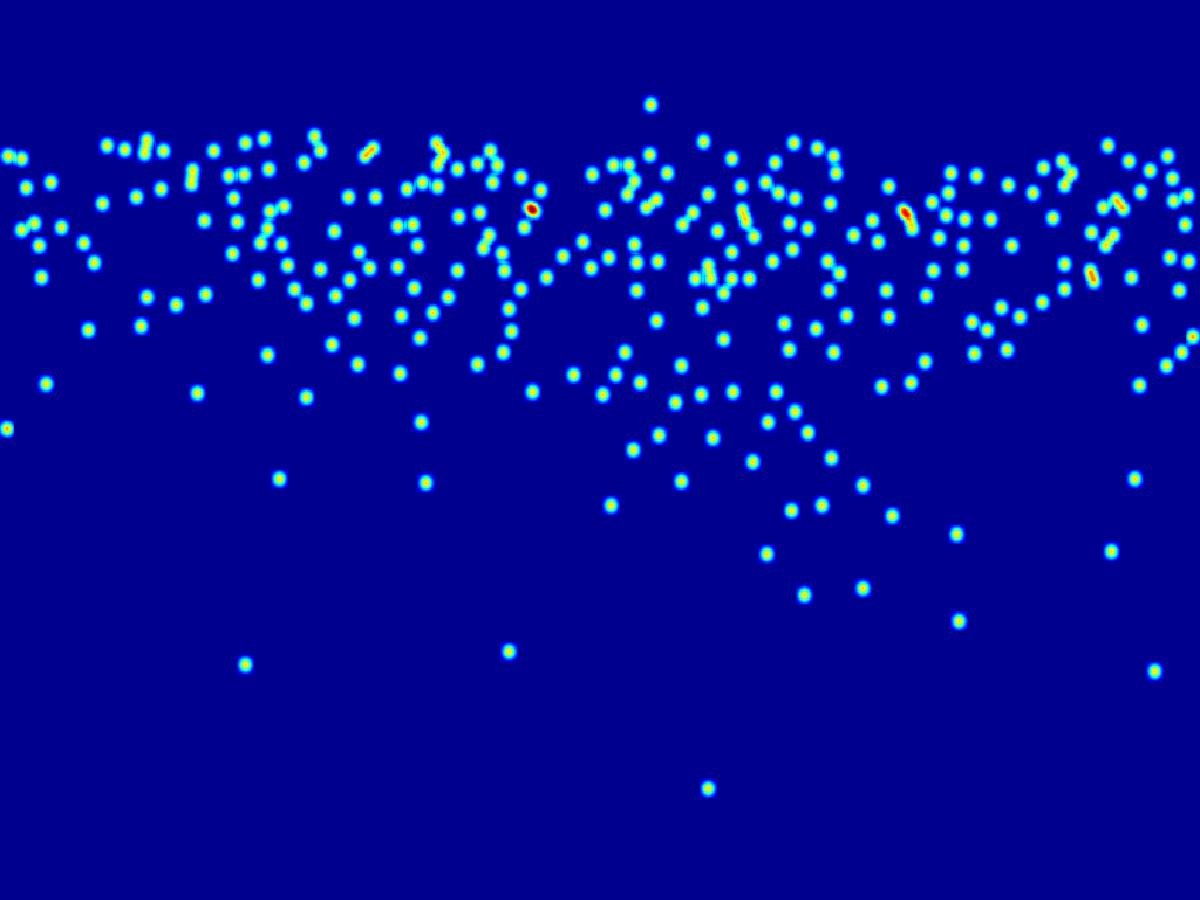}
     }
     &
     {\includegraphics[width=0.22\linewidth]{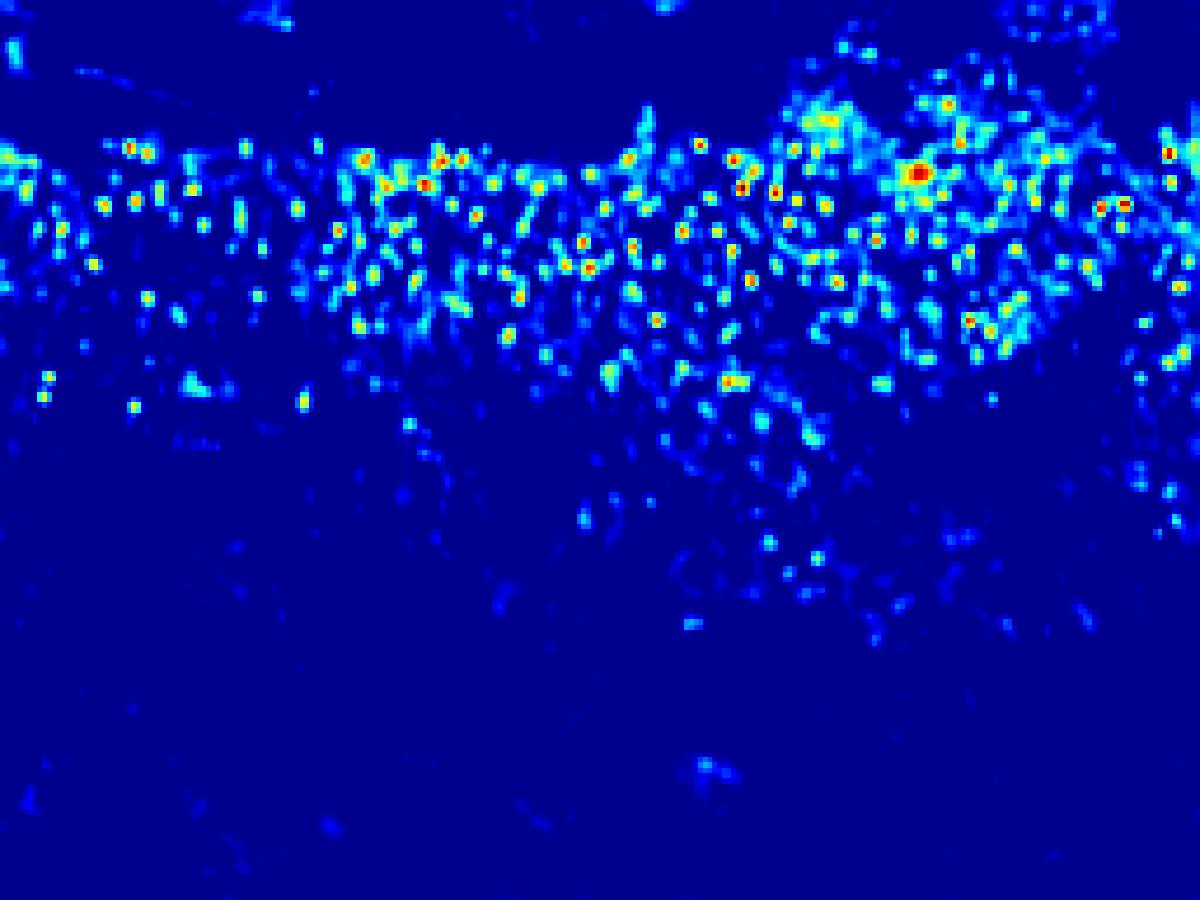}
     } & 
     {\includegraphics[width=0.22\linewidth]{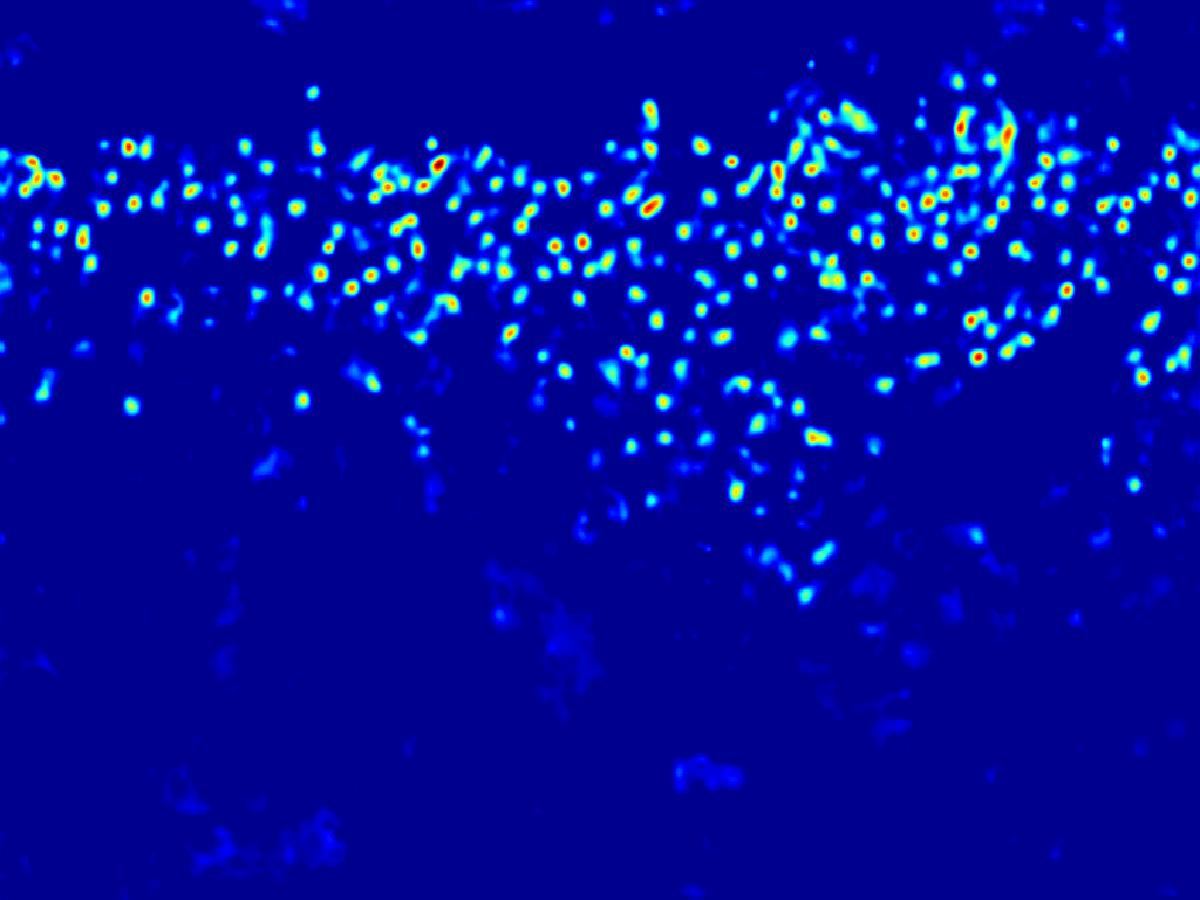}
     } \\
    &270  & 346 & 280\\
   \\ 
{  
	\includegraphics[width=0.22\linewidth]{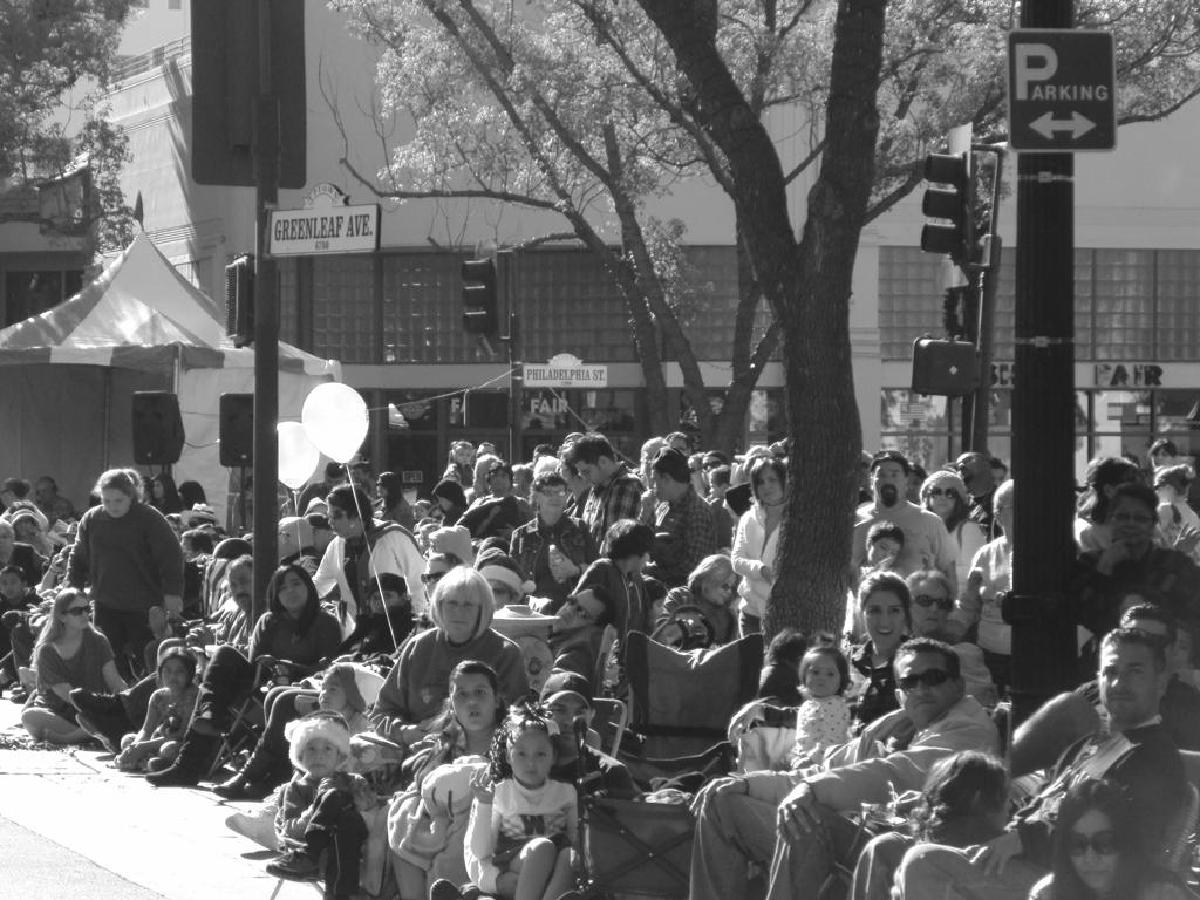}
     } &  {  
	\includegraphics[width=0.22\linewidth]{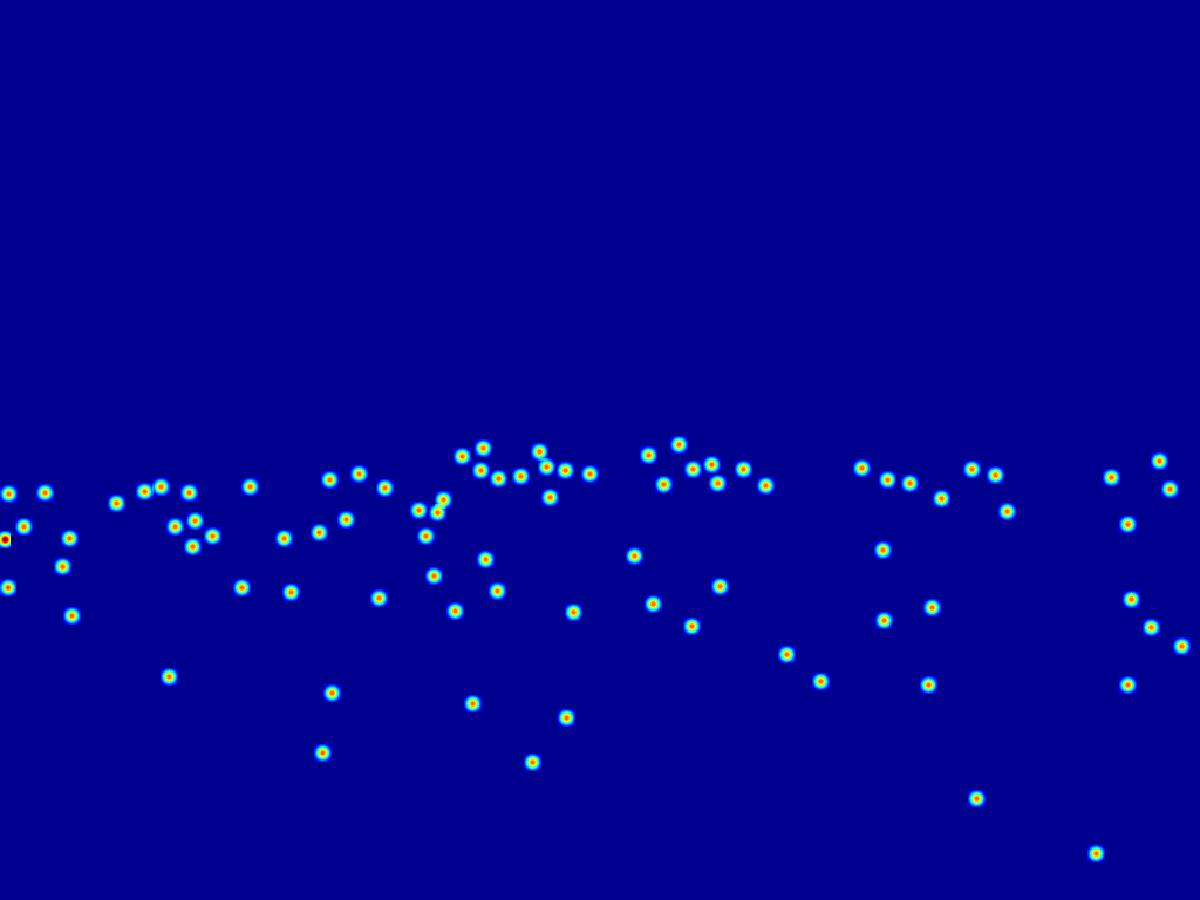}
     }
     &
     {\includegraphics[width=0.22\linewidth]{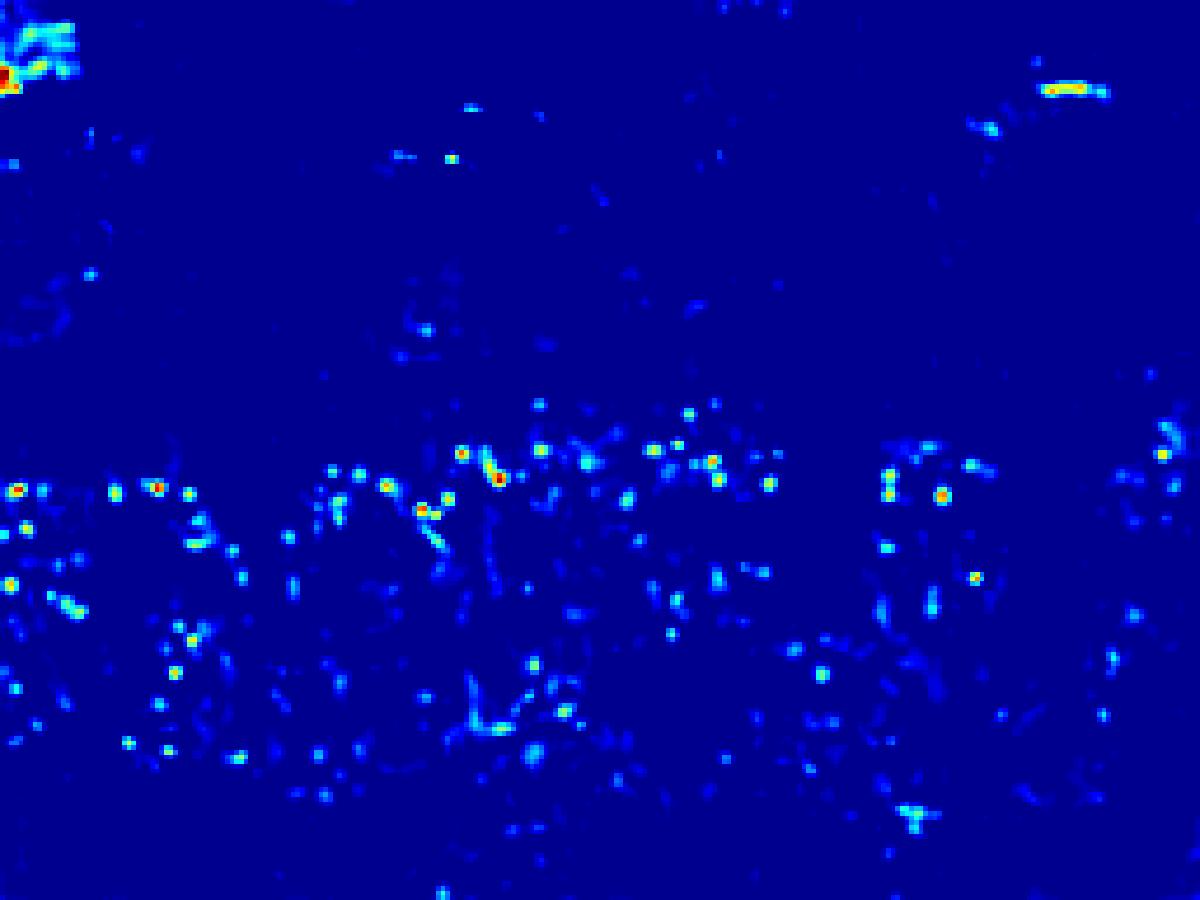}
     } & 
     {\includegraphics[width=0.22\linewidth]{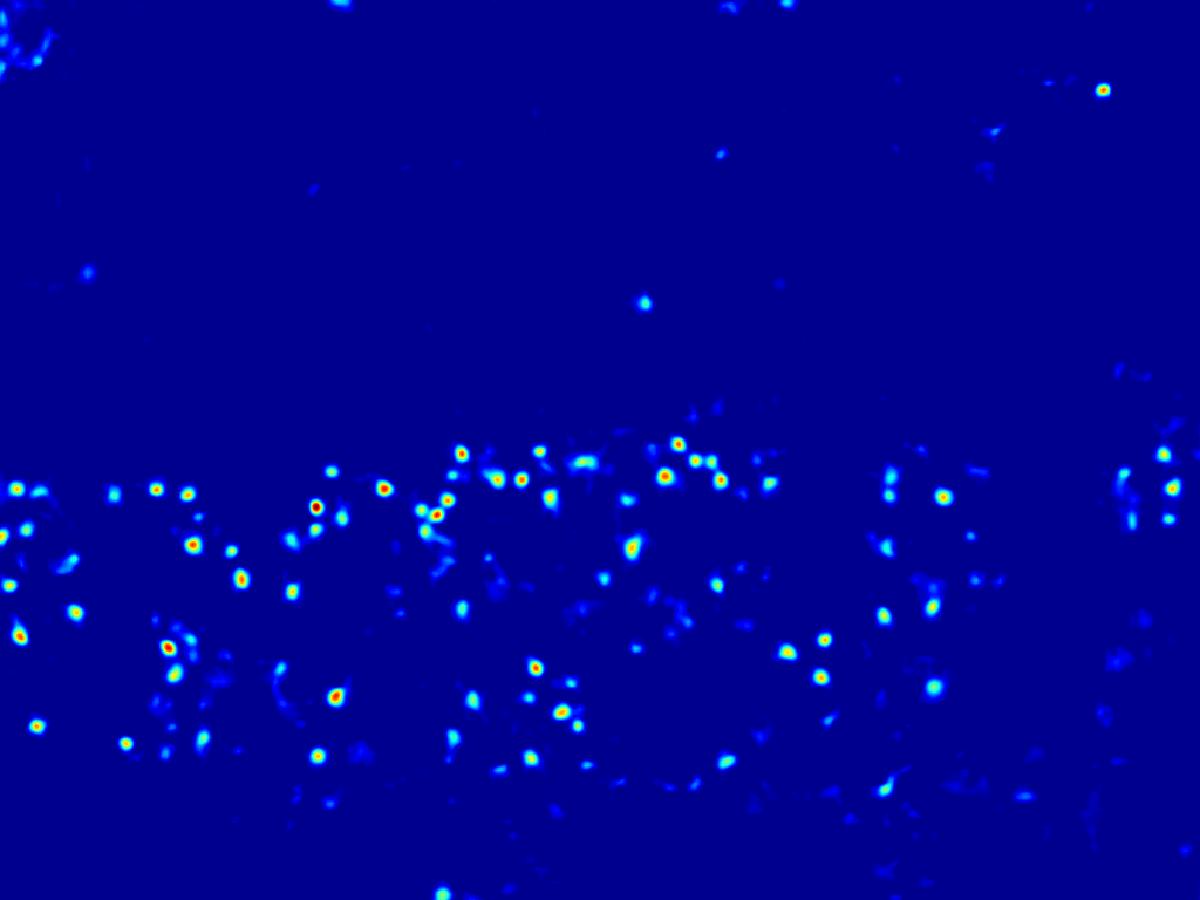}
     } \\
    &86  & 114 & 89\\
   \\ 
   {  
	\includegraphics[width=0.22\linewidth,height=0.22\linewidth]{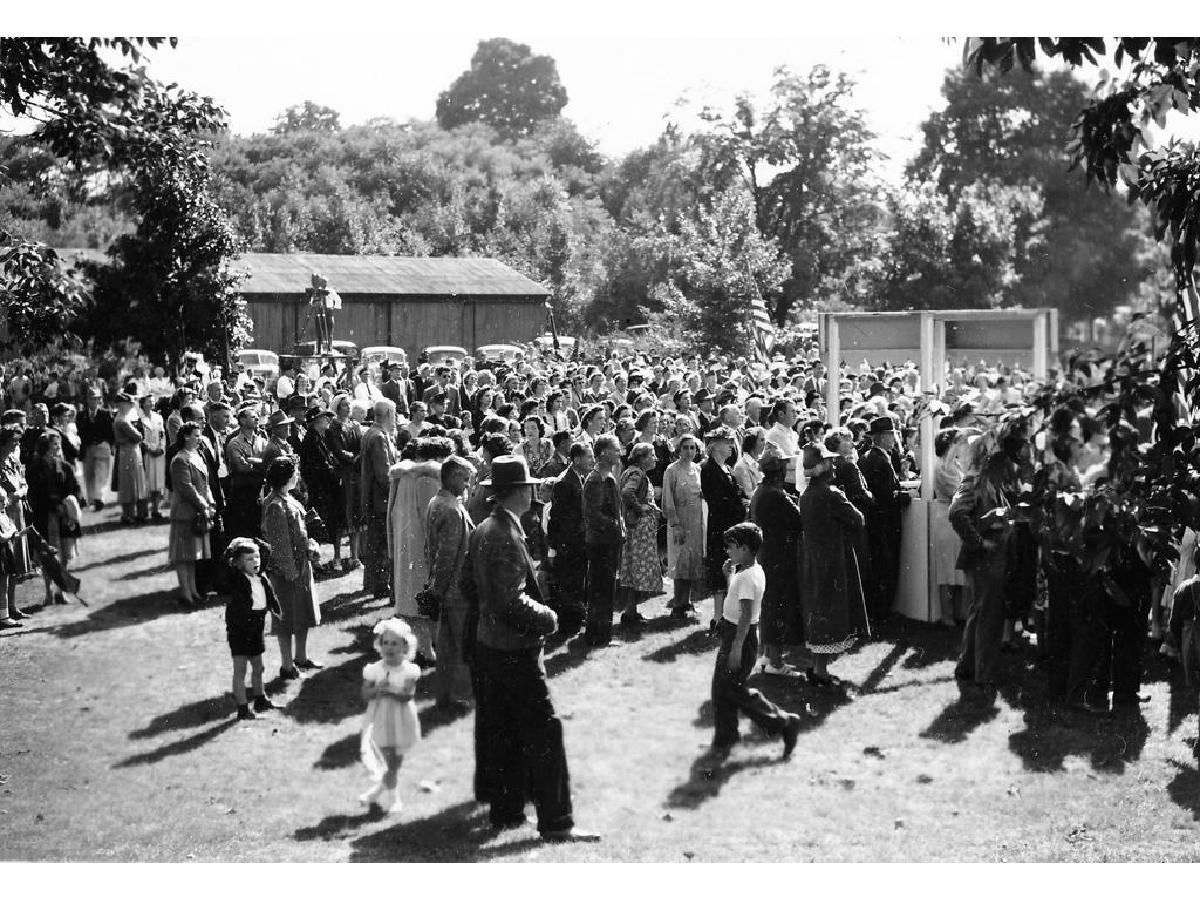}
     } &  {  
	\includegraphics[width=0.22\linewidth,height=0.22\linewidth]{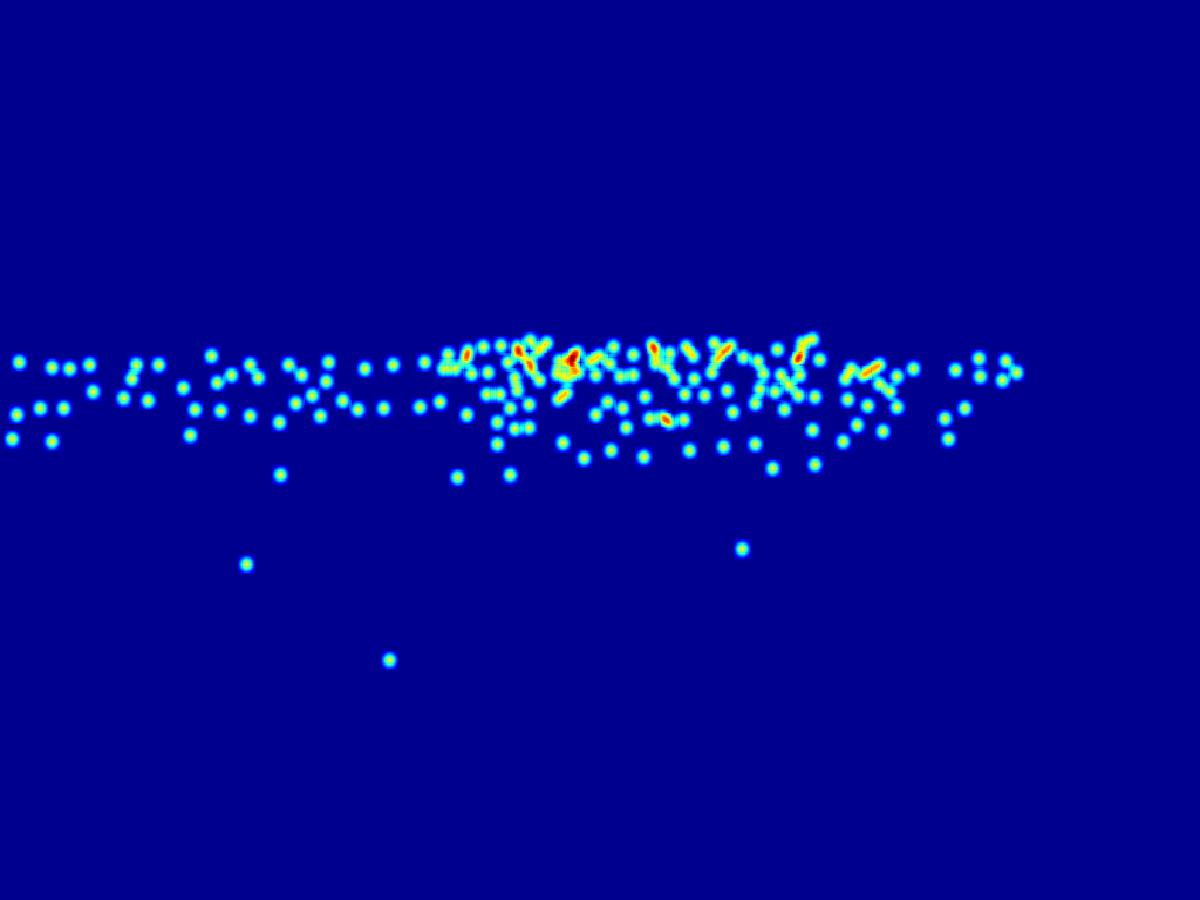}
     }
     &
     {\includegraphics[width=0.22\linewidth,height=0.22\linewidth]{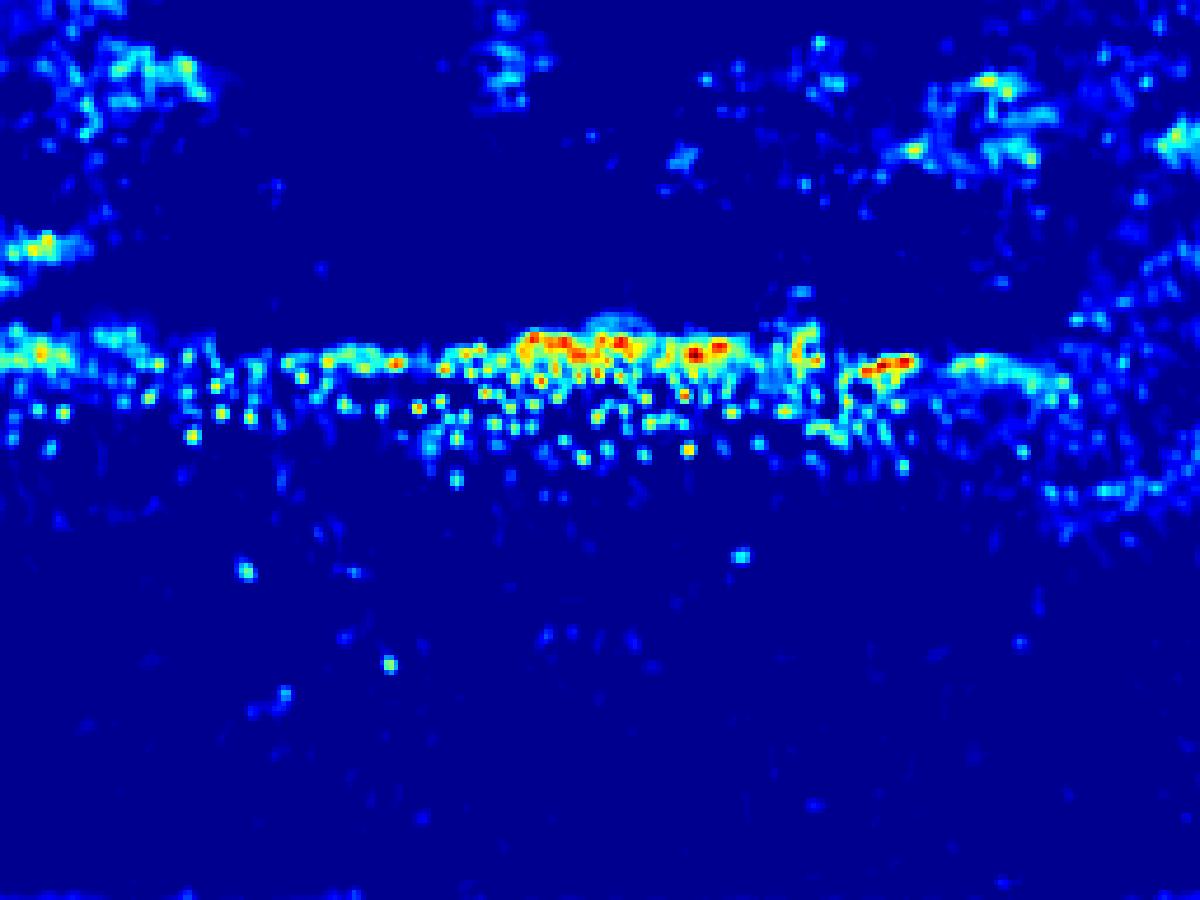}
     } & 
     {\includegraphics[width=0.22\linewidth,height=0.22\linewidth]{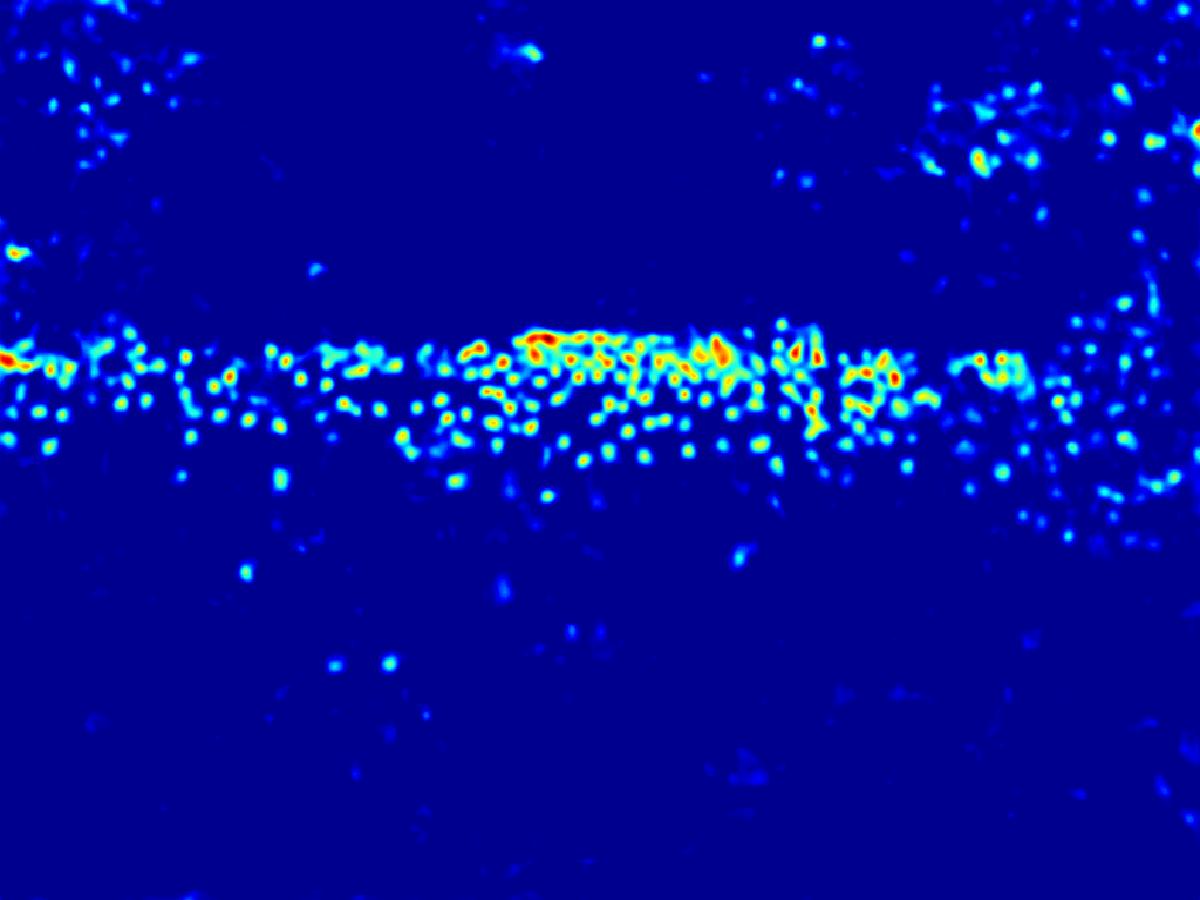}
     } \\
    &172  & 493 & 317\\
   \\ 
   {  
	\includegraphics[width=0.22\linewidth]{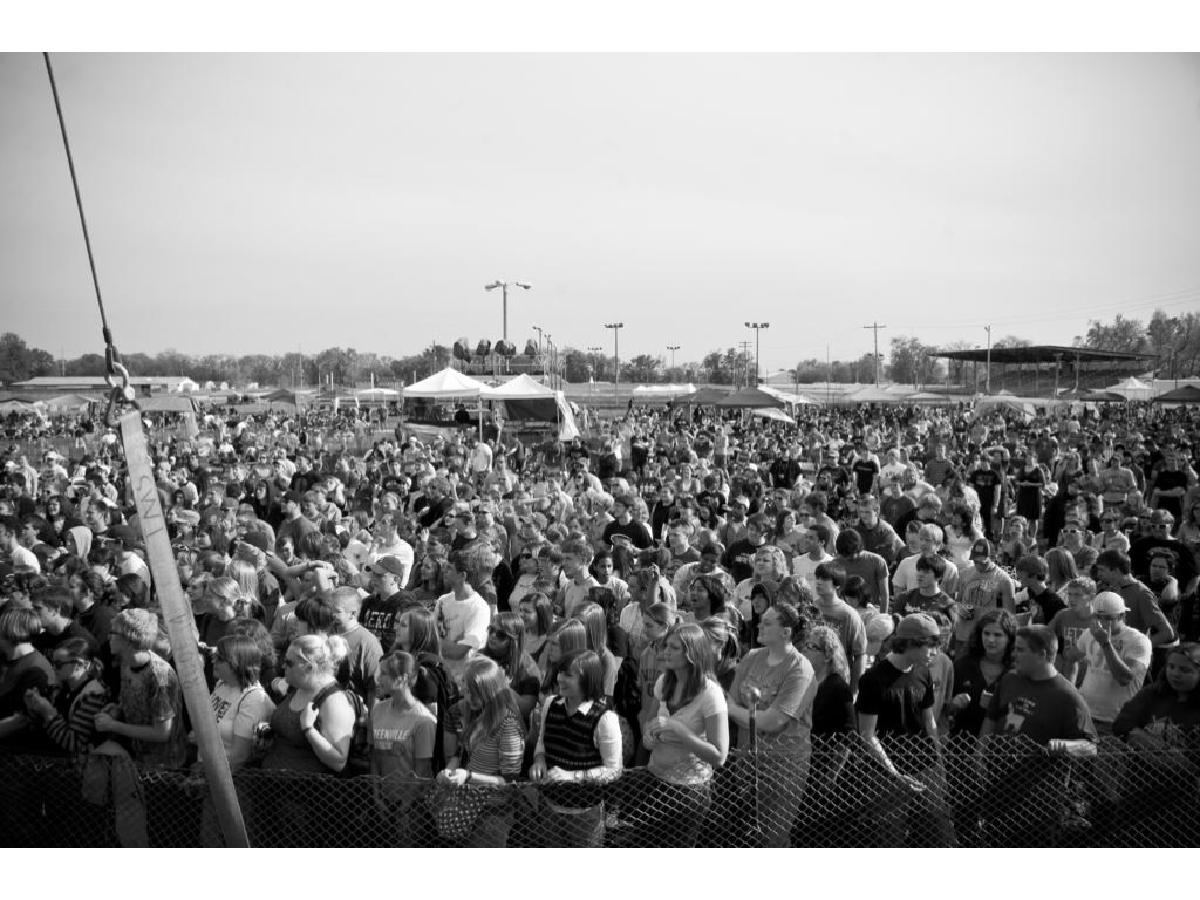}
     } &  {  
	\includegraphics[width=0.22\linewidth]{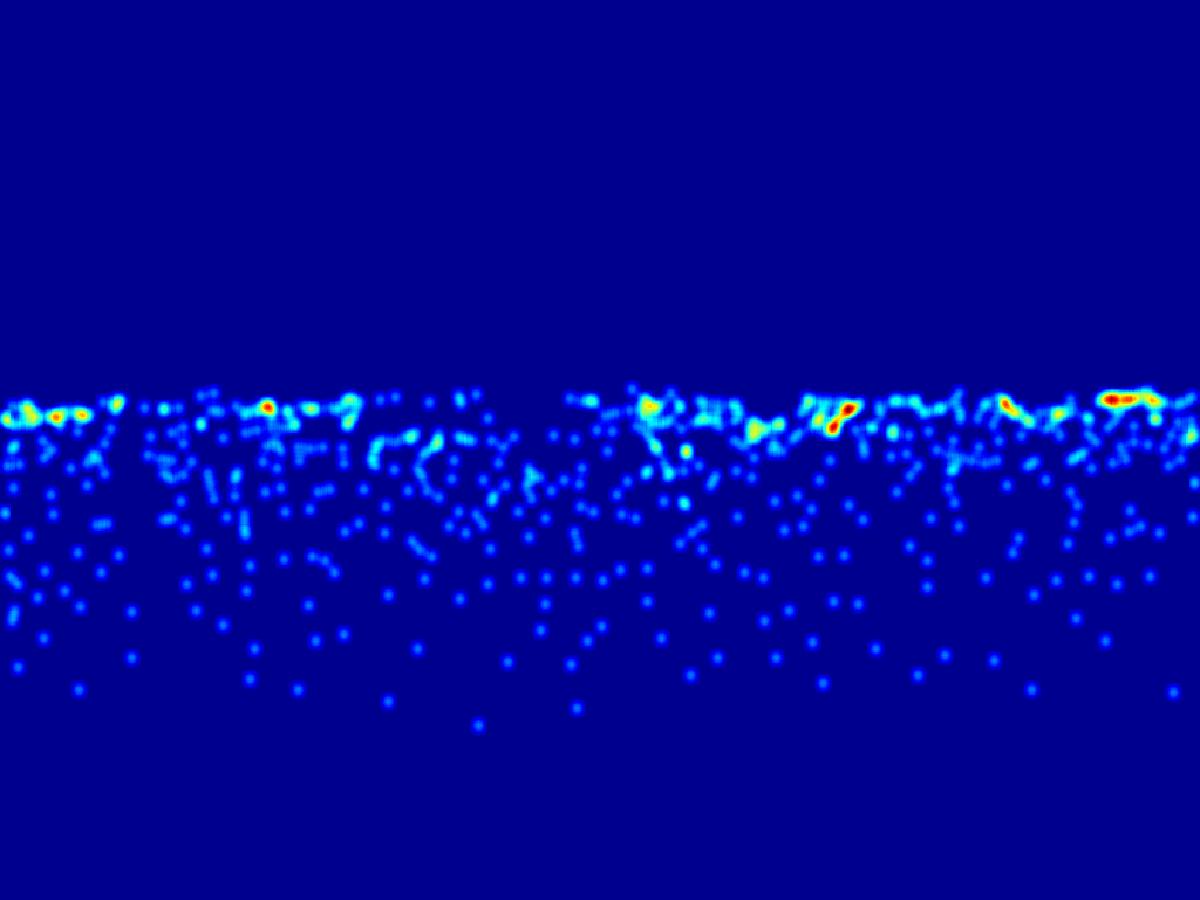}
     }
     &
     {\includegraphics[width=0.22\linewidth]{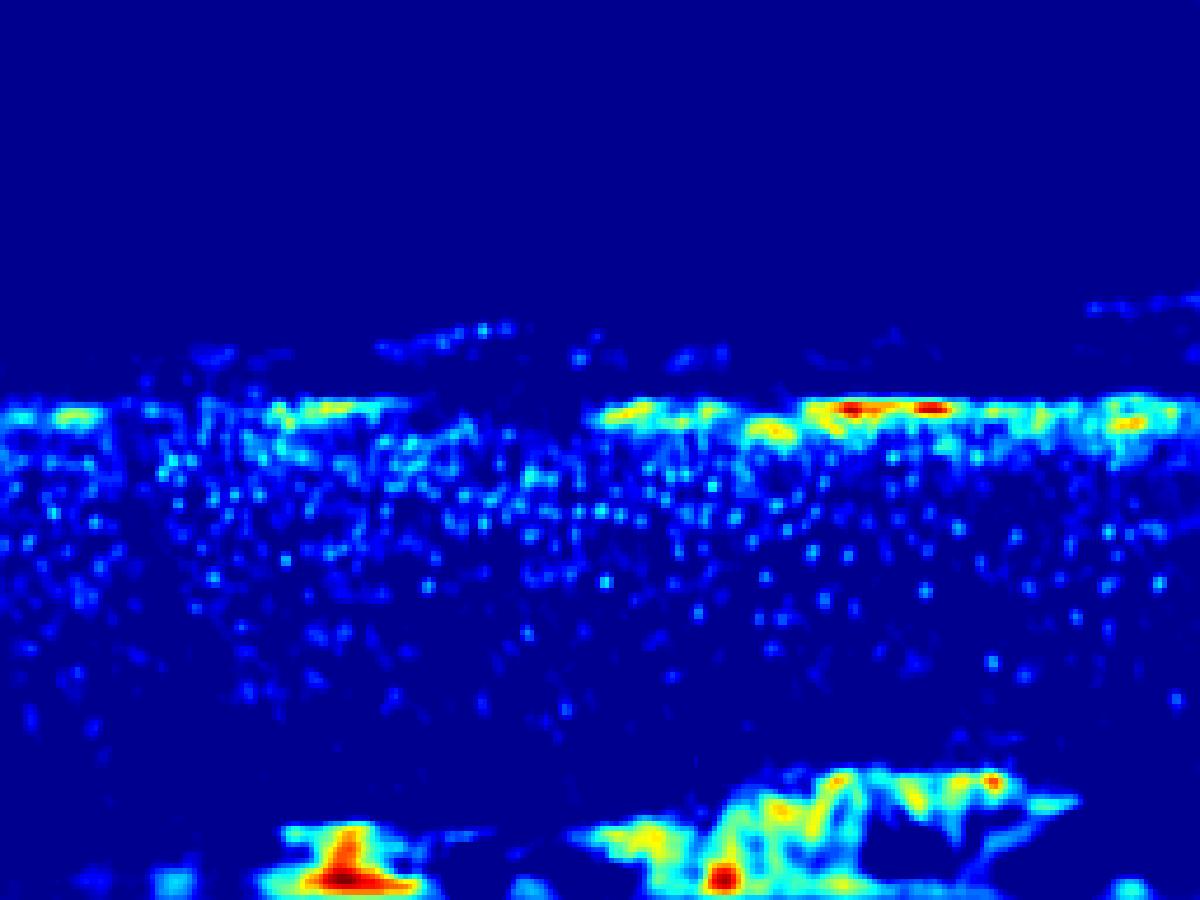}
     } & 
     {\includegraphics[width=0.22\linewidth]{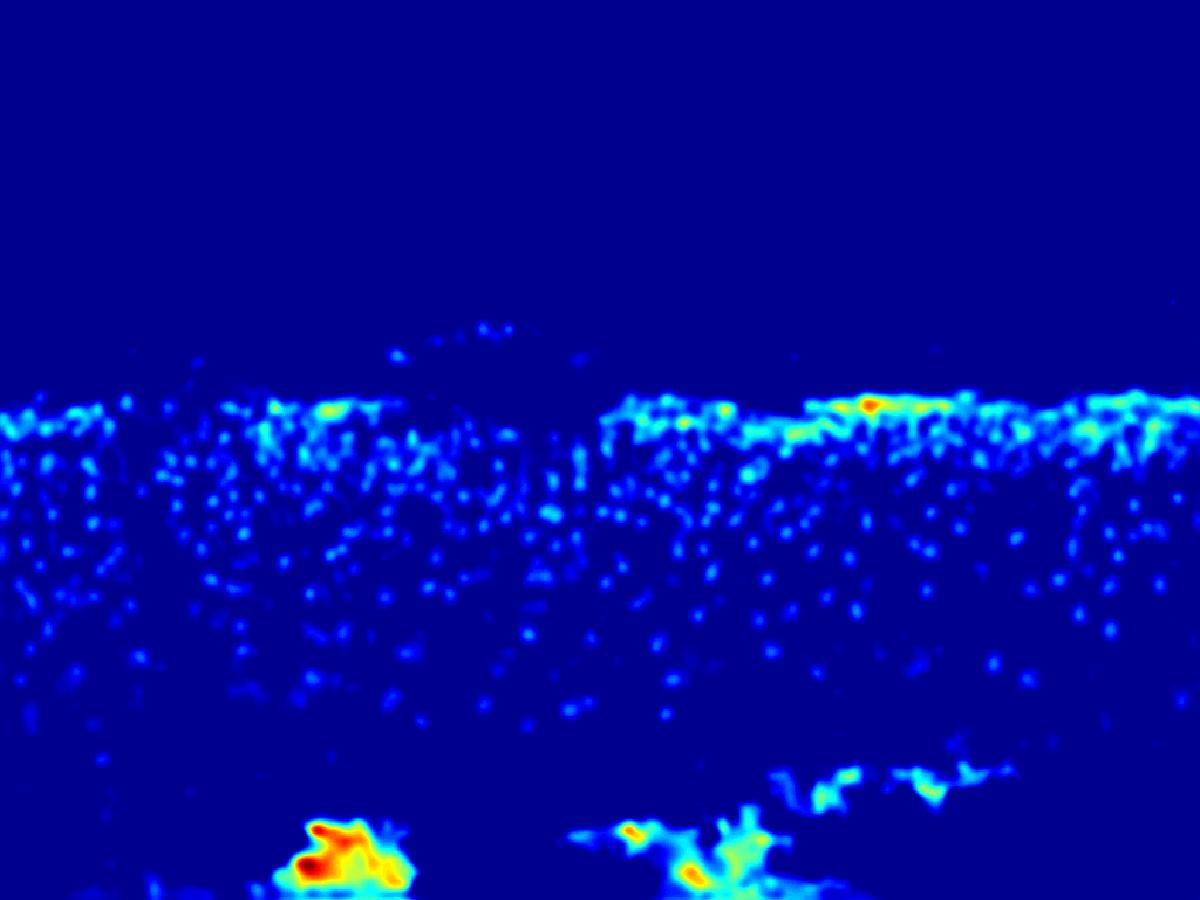}
     } \\
    &566  & 961 & 744\\
\end{tabular}
\caption{{\bf Qualitative results, some success and failure cases.} The four columns show the input image, ground truth annotation map, the low resolution prediction (LR output), and the high resolution prediction map (HR output). The total counts are shown below each density map. The first three rows are success cases for ic-CNN, while the last two are failure cases. ic-CNN sometimes misclassifies tree leaves as people.\label{quali}}
\end{figure}

\begin{figure}[!thb]
\begin{tabular}{cccc} 

{\bf Image} & {\bf Ground truth} & {\bf LR output} & {\bf HR  output} \\ \\ 
{  
	\includegraphics[width=0.22\linewidth]{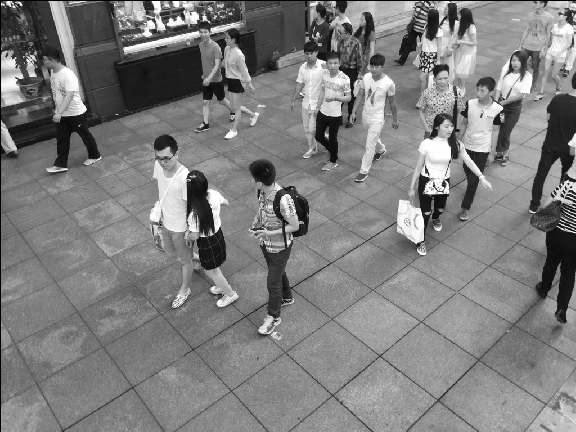}
     } &  {  
	\includegraphics[width=0.22\linewidth]{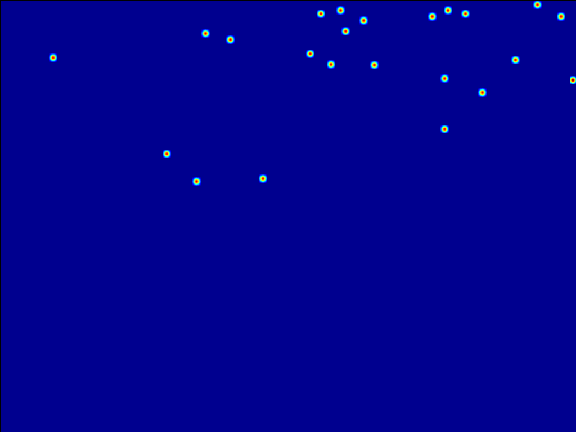}
     }
     &
     {\includegraphics[width=0.22\linewidth,]{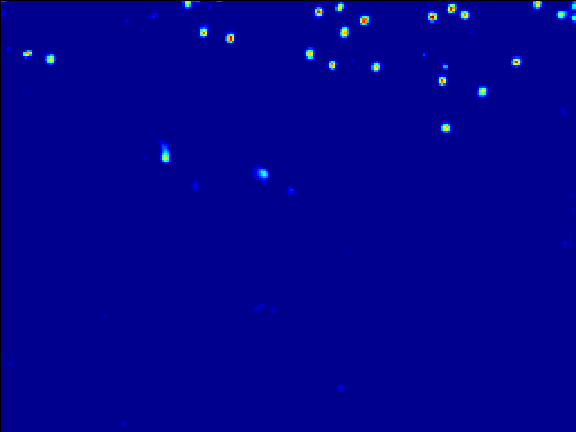}
     } & 
     {\includegraphics[width=0.22\linewidth]{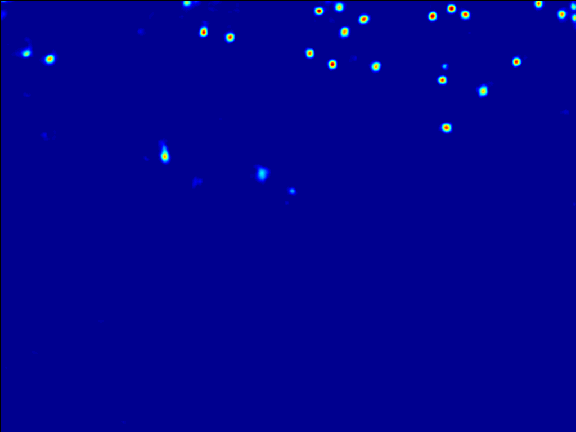}
     } \\
    &23  & 26 & 24\\
   \\ 

{  
	\includegraphics[width=0.22\linewidth]{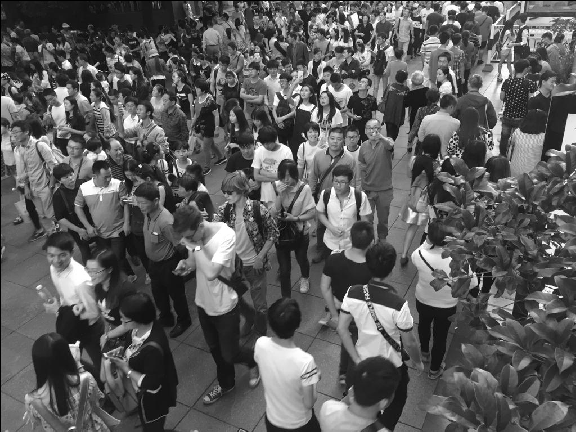}
     } &  {  
	\includegraphics[width=0.22\linewidth]{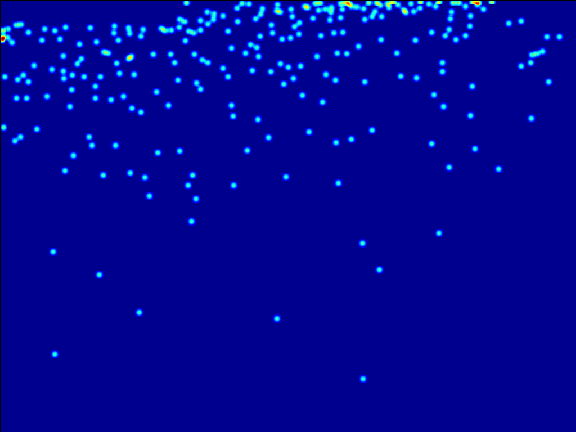}
     }
     &
     {\includegraphics[width=0.22\linewidth]{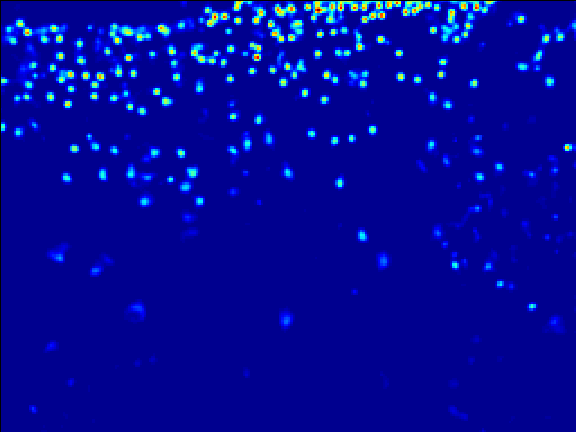}
     } & 
     {\includegraphics[width=0.22\linewidth]{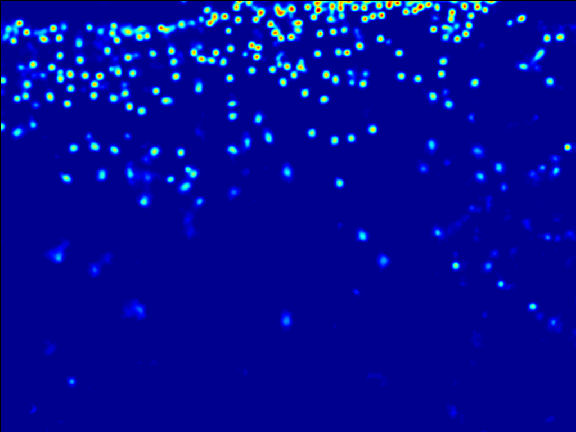}
     } \\
    &252  & 257 & 252\\
   \\ 

{  
	\includegraphics[width=0.22\linewidth]{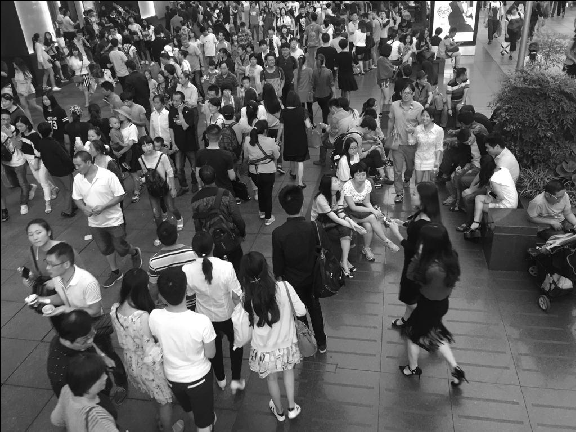}
     } &  {  
	\includegraphics[width=0.22\linewidth]{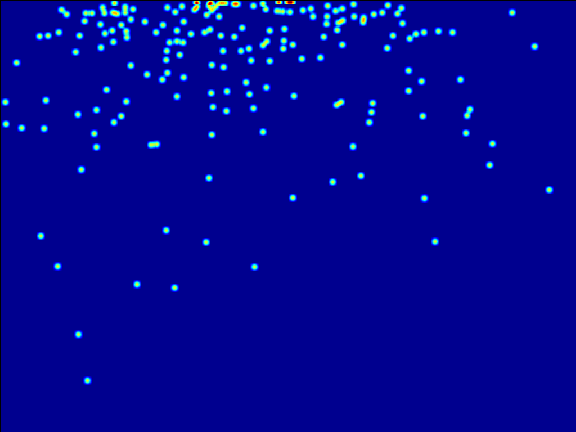}
     }
     &
     {\includegraphics[width=0.22\linewidth]{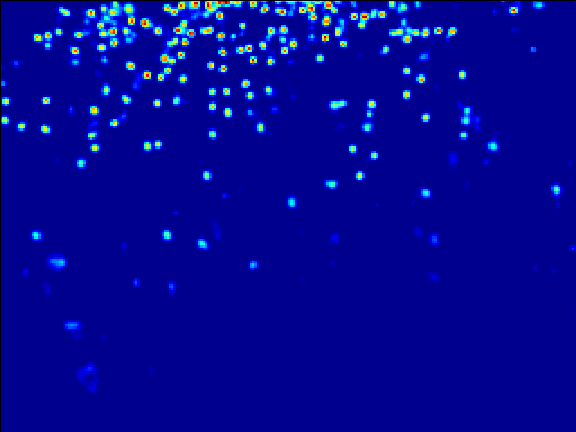}
     } & 
     {\includegraphics[width=0.22\linewidth]{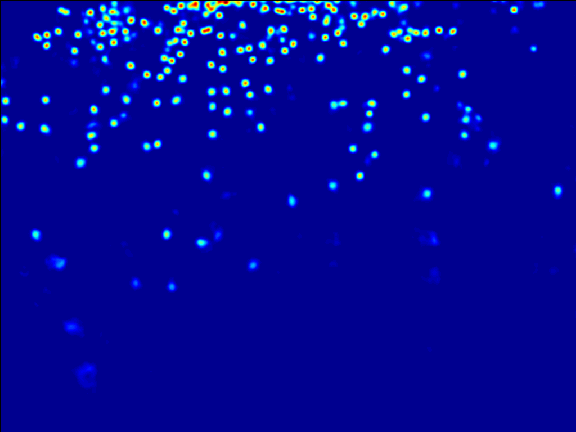}
     } \\
    &183  & 191 &186\\
   \\ 
   {  
	\includegraphics[width=0.22\linewidth]{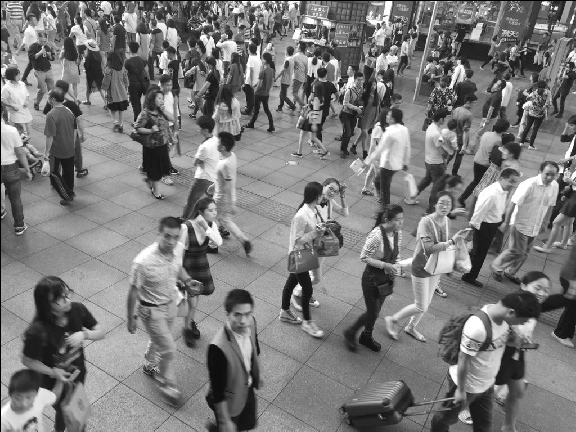}
     } &  {  
	\includegraphics[width=0.22\linewidth]{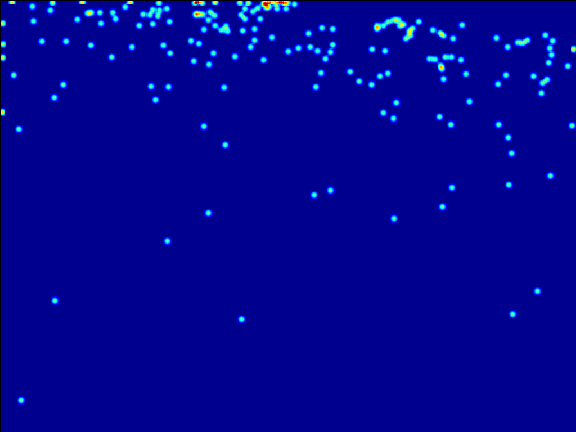}
     }
     &
     {\includegraphics[width=0.22\linewidth]{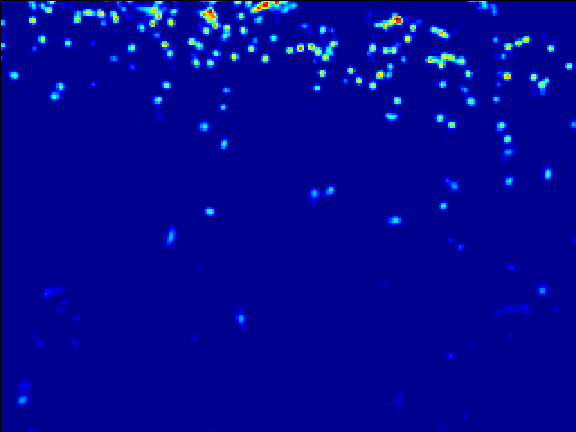}
     } & 
     {\includegraphics[width=0.22\linewidth]{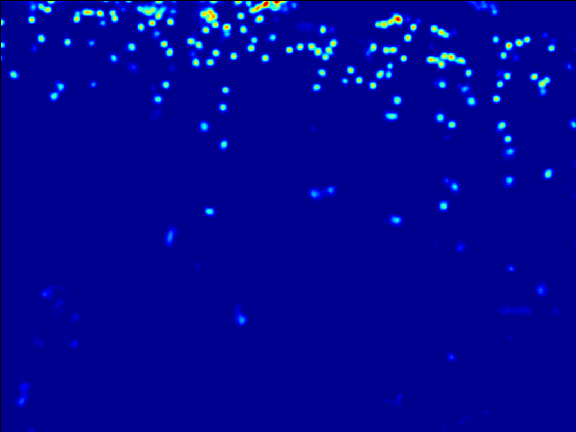}
     } \\
    &181  & 167 & 164\\
   \\

   {  
	\includegraphics[width=0.22\linewidth]{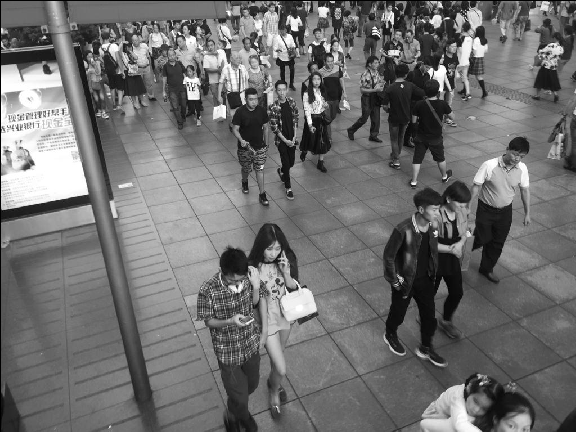}
     } &  {  
	\includegraphics[width=0.22\linewidth]{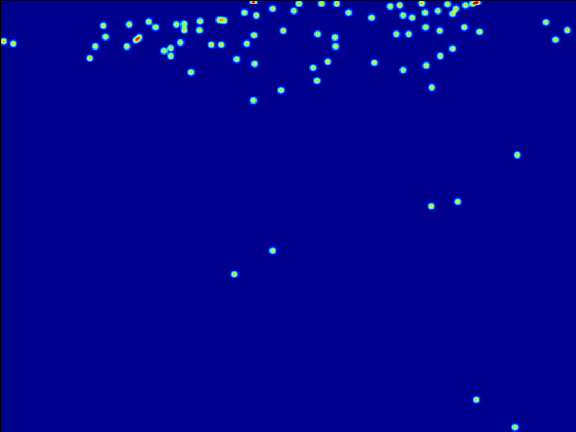} 
     }
     &
     {\includegraphics[width=0.22\linewidth]{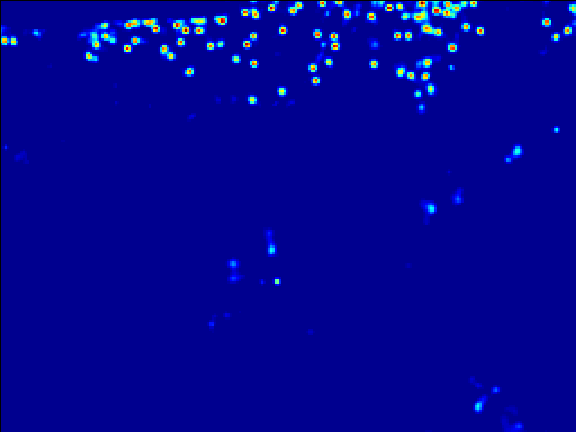}
     } & 
     {\includegraphics[width=0.22\linewidth]{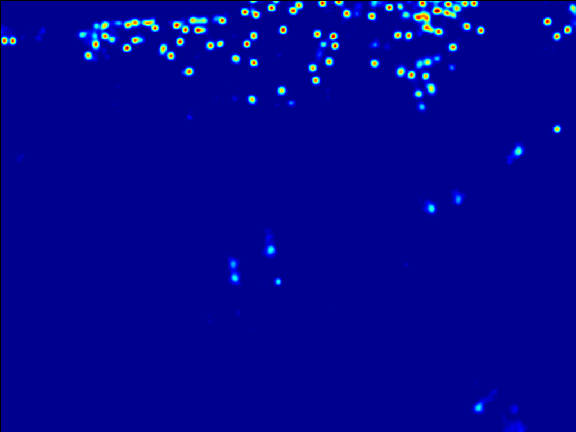}
     } \\
    &84 & 109 & 103\\
\end{tabular}
\caption{{\bf Qualitative results on the Shanghaitech Part B dataset.} The four columns show the input image, the ground truth annotation map, the low resolution prediction (LR output), and the high resolution prediction map (HR output). Underneath each density map is the total count, rounded to the nearest integer.\label{quali2}}
\vskip -0.2in
\end{figure}

In Figure~\ref{quali}, we show some qualitative results on images from the Shanghaitech Part-A dataset obtained using ic-CNN. The first three are success cases for ic-CNN, while the last two are failure cases. In the failure cases, we see that ic-CNN sometimes misclassify tree leaves as tiny people in a crowd. In Figure~\ref{quali2}, we show some qualitative results on images from Shanghaitech Part-B dataset.

\section{Conclusions}
In this paper, we have proposed ic-CNN, a two-branch architecture for crowd counting via crowd density estimation based. We have also proposed a multi-stage pipeline comprising of multiple ic-CNNs, where each stage takes into account the predictions of all the previous stages. We performed experiments on three challenging crowd counting benchmark datasets and observed the effectiveness of our iterative approach.


\setlength{\tabcolsep}{3pt}

\setlength{\tabcolsep}{3pt}

\myheading{Acknowledgements.} This work was supported by SUNY2020 Infrastructure Transportation Security Center. The authors would like to thank Boyu Wang for participating on the discussions and experiments related to an earlier version of the proposed technique. The authors would like to thank NVIDIA for their GPU donation. 

\bibliographystyle{splncs04}
\bibliography{longstrings,eccvbib,pubs}
\end{document}